%% file: main.tex
\documentclass[10pt,twocolumn,letterpaper]{article}

\usepackage[pagenumbers]{iccv} % To force page numbers, e.g. for an arXiv version
\usepackage{float}

\input{preamble}

\definecolor{iccvblue}{rgb}{0.21,0.49,0.74}
\usepackage[pagebackref,breaklinks,colorlinks,allcolors=iccvblue]{hyperref}

\def\paperID{15119} % *** Enter the Paper ID here
\def\confName{ICCV}
\def\confYear{2025}

\title{\includegraphics[height=1em]{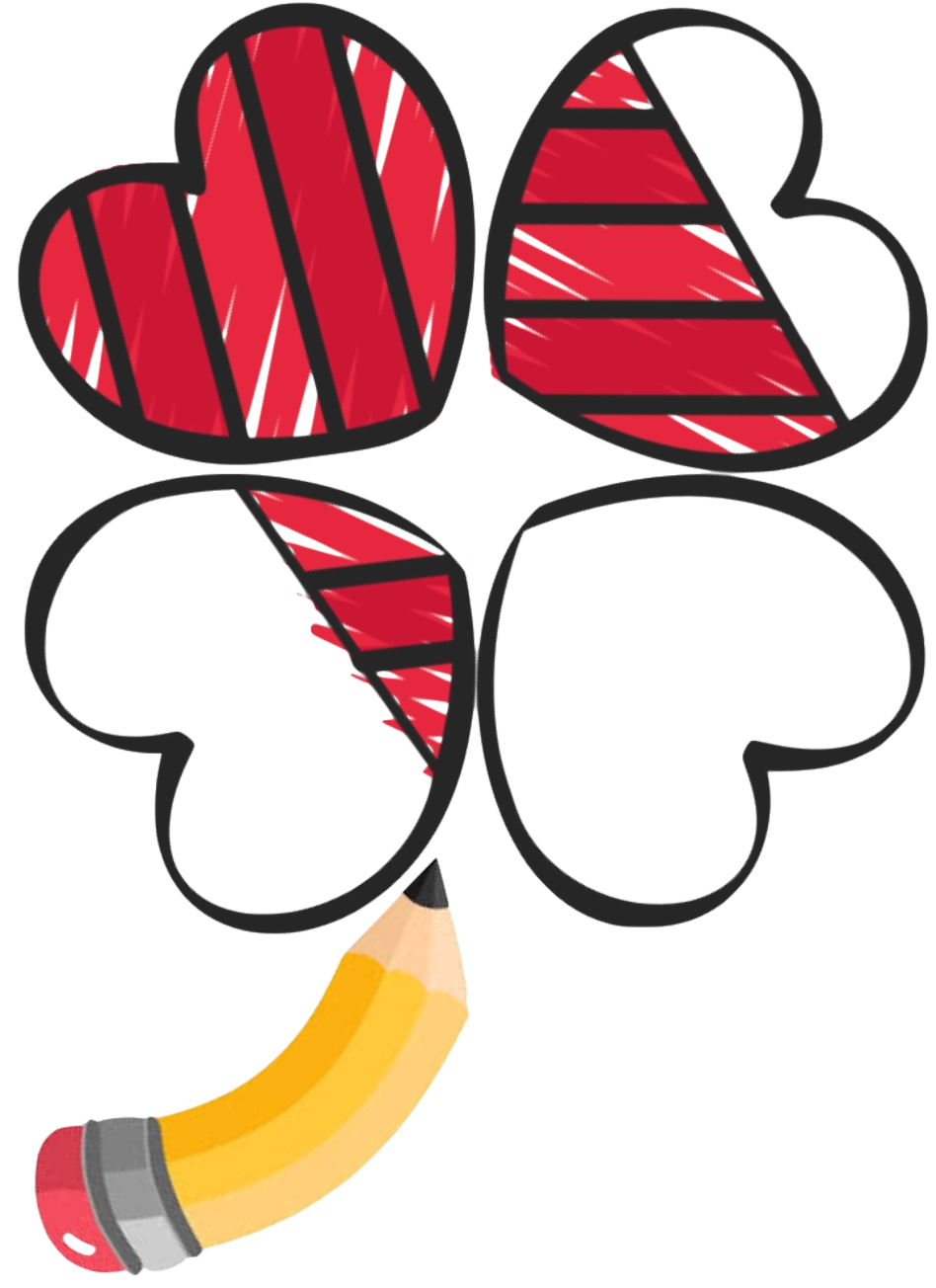}
Moodifier: MLLM-Enhanced Emotion-Driven Image Editing
}

\author{Jiarong Ye\\
The Pennsylvania State University\\
{\tt\small jxy225@psu.edu}
\and
Sharon X. Huang\\
The Pennsylvania State University\\
{\tt\small suh972@psu.edu}
}

\begin{document}
\maketitle
\input{sec/0_abstract}    
\input{sec/1_intro}
\input{sec/2_related_works}
\input{sec/3_method}

\input{sec/4_experiments}
\input{sec/5_conclusion}
{
    \small
    \bibliographystyle{unsrt}
    \bibliography{main}
}

\end{document}

%% file: preamble.tex
\usepackage{algorithm}
\usepackage{algorithmic} 
\usepackage{bbm}
\usepackage{multirow}

%% file: sec/0_abstract.tex
\begin{abstract}

%Emotional control in visual content has promising application potential across creative industries, yet precise manipulation remains challenging because emotions are subjective, abstract, and manifest differently across contexts.
Bridging emotions and visual content for emotion-driven image editing holds great potential in creative industries, yet precise manipulation remains challenging due to the abstract nature of emotions and their varied manifestations across different contexts. We tackle this challenge with an integrated approach consisting of three complementary components. 
First, we introduce MoodArchive, an 8M+ image dataset with detailed hierarchical emotional annotations generated by LLaVA and partially validated by human evaluators. Second, we develop MoodifyCLIP, a vision-language model fine-tuned on MoodArchive to translate abstract emotions into specific visual attributes. Third, we propose Moodifier, a training-free editing model leveraging MoodifyCLIP and multimodal large language models (MLLMs) to enable precise emotional transformations while preserving content integrity. Our system works across diverse domains such as character expressions, fashion design, jewelry, and home décor, enabling creators to quickly visualize emotional variations while preserving identity and structure. Extensive experimental evaluations show that Moodifier outperforms existing methods in both emotional accuracy and content preservation, providing contextually appropriate edits. By linking abstract emotions to concrete visual changes, our solution unlocks new possibilities for emotional content creation in real-world applications. 
% We will release the MoodArchive dataset, MoodifyCLIP model, and make the Moodifier code and demo publicly available upon acceptance.
Project website: \href{https://moodify2024.github.io/app/}{this https URL}.

% Focus:
%     (1. Moodarchive that with 8m+ image-text pairs, diverse emotional ranges and human-verified caps generated by mllm
%     2. MoodifyClip that enhanced to deal with long detailed cap with emotional elements)
%     3. Moodifier that as an emotion-driven smart editing system with an interactive tool to accommodates flexible user requests, addressing the subjective nature of emotions where auto-determined changes may not always align with user preferences

% Selling points:
%     1. MoodifyClip's plug-and-play functionality enables direct replacement of OpenAI CLIP in existing market pipelines such as StableDiffusion
%     2. Zero pre-training requirements for Moodifier, leveraging emotional semantic understanding from both multi-modal language models and MoodifyClip
%     3. Interactive tool for subjective emotional adjustments, all users need to do is to input emotion
    
\end{abstract}

%% file: sec/1_intro.tex
\vspace{-10pt}
\section{Introduction}
\label{sec:intro}

%-------------------------------------------------------------------------

\begin{figure*}[htbp]
    \centering
    \includegraphics[width=0.94\textwidth]{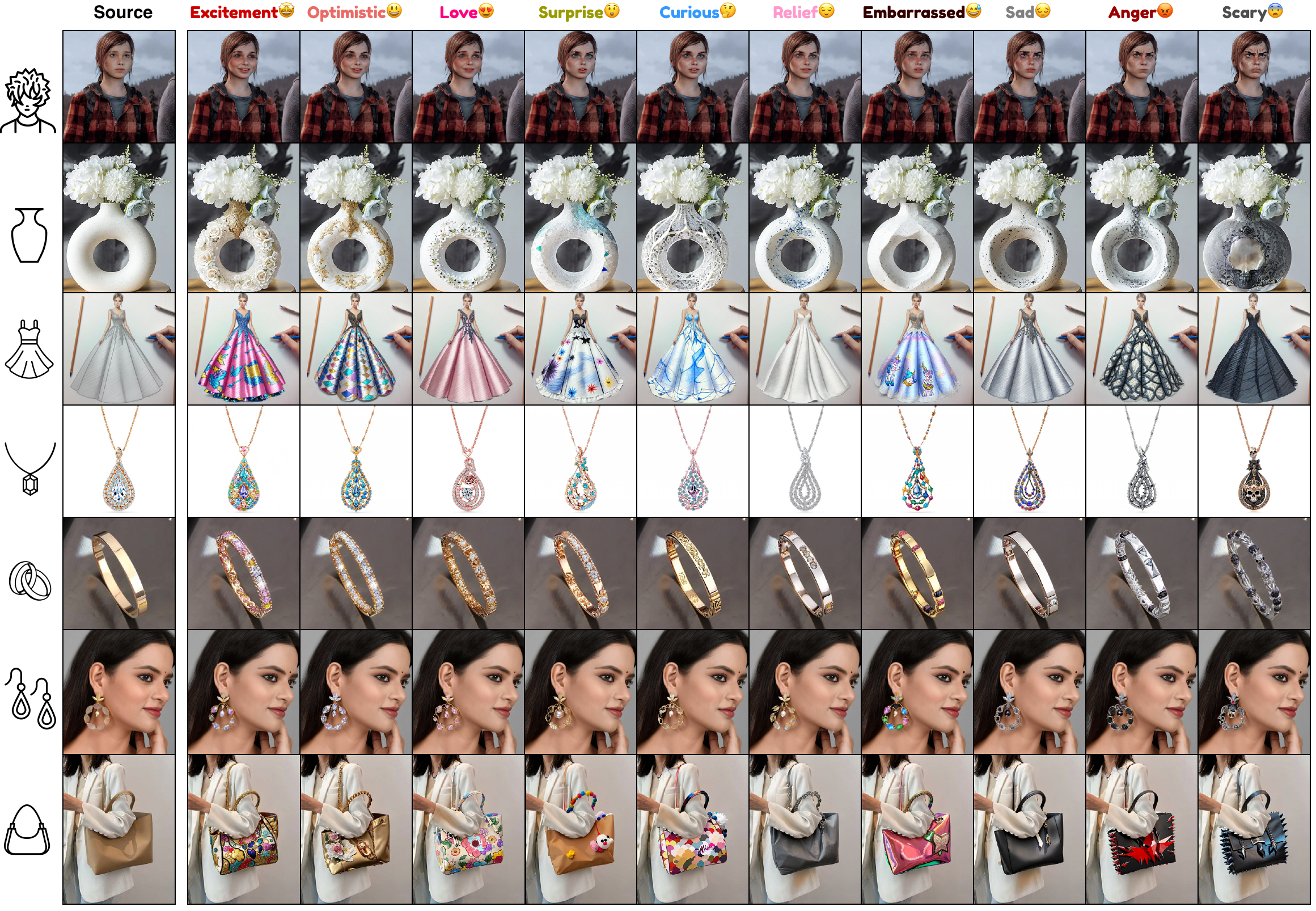}
    \caption{Our emotion-driven Moodifier image editing system performs consistently across diverse scenarios such as product mockups, e-commerce, fashion, jewelry, and home décor. It precisely modifies target objects while preserving the overall structure and integrity of the scene. 
    %preserves unedited elements while precisely modifying targets. 
    The target objects selected for editing are depicted as clip arts in the leftmost column. In non-facial contexts, emotions manifest as semantic qualities (e.g., ``Excitement" as vibrant designs, ``Sad" as drooping elements, ``Anger" as bold styling). Zoom in for better view.}
\label{fig:moodifier_examples}
\vspace{-14pt}
\end{figure*}

Imagine transforming a cocktail dress from elegant to joyful with a single click, or visualizing how a pendant might evoke different emotions through subtle material changes. As demonstrated in Fig.  \ref{fig:moodifier_examples}, tools for precise emotional manipulation could significantly enhance creative workflows\footnote{A demo video of Moodifier in action is available in the project site. Please go check it out.}.
Beyond aesthetic exploration, such capabilities would create tangible value across domains: animation artists could generate emotional sequences while preserving character identity; e-commerce platforms could tailor product imagery to different customer segments; marketing teams could repurpose assets across emotional campaigns without additional photoshoots, reducing production time and costs. 

However, existing image editing tools lack sophisticated emotion-driven controls. The challenge lies in how emotions manifest differently across contexts; for example, ``excitement'' translates to vibrant patterns in fashion, specific facial expressions in characters, and intricate textures in jewelry. Existing methods (e.g. \cite{Zhang2023AddingCC, Brooks2022InstructPix2PixLT, Geng23instructdiff}) either provide simplistic adjustments, such as generic filters, or introduce unwanted artifacts that compromise design integrity (see Fig. \ref{fig:moodifier_comparisons}),
forcing creators to rely on time-consuming iterations rather than enabling seamless creative exploration.

Here we propose Moodifier to address previous limitations by bridging abstract emotional concepts with precise visual modifications through a two-stage approach, as illustrated in Fig.  \ref{fig:moodifer_workflow}. First, given a desired emotion, MLLMs analyze the source image to generate context-specific editing instructions and spatial attention maps, highlighting where emotional modifications should be applied.
To translate these instructions into visual changes, Moodifier uses MoodifyCLIP, a vision-language model fine-tuned specifically for emotional understanding. Unlike standard CLIP models trained on brief, factual captions from datasets like COCO~\cite{chen2015microsoft} and Flickr~\cite{young-etal-2014-image}), MoodifyCLIP utilizes our MoodArchive dataset, which contains millions of affective images (See Fig.  1 in supp.) with hierarchical emotional annotations as shown in Fig.  \ref{fig:moodifyclip_recap}.

Second, using MoodifyCLIP-processed prompts and spatial attention maps, Moodifier performs attention-controlled editing. It extracts the latent representation of the source image and directs the diffusion model to modify only emotion-relevant attributes while preserving the overall content. The visual results in Fig.  \ref{fig:moodifier_examples} demonstrate Moodifier's versatility: character faces express diverse emotions while maintaining identity, dresses adapt to different emotional tones while retaining their original design, and jewelry and décor shift in mood without altering their fundamental structure.

To summarize, we introduce a comprehensive framework that advances emotion-driven image editing at the intersection of computer vision and affective computing:
%present an integrated suite of tools for emotional visual editing:

\noindent \textbf{1. MoodArchive:} The largest dataset of its kind, MoodArchive consists of over 8 million images with 
%An 8M+ image dataset with 
hierarchical emotional annotations generated by LLaVA-NeXT and partially validated by human evaluators. 
This dataset establishes a new benchmark for training models to understand nuanced emotional attributes in visual content. 
\\
% providing the essential foundation for nuanced emotional understanding.\\
\noindent \textbf{2. MoodifyCLIP:} A vision-language model fine-tuned on MoodArchive that extends the capabilities of existing CLIP models by enabling fine-grained emotional reasoning and translation of abstract emotions into specific, context-aware visual attributes.\\
%Built on MoodArchive, this vision-language model learns to translate abstract emotions into specific visual attributes.
\textbf{3. Moodifier:} A novel training-free image editing system that seamlessly integrates MoodifyCLIP's emotional intelligence with MLLMs and diffusion models. Moodifier enables precise, targeted emotional transformations while preserving structural and semantic integrity. 
%Leveraging MoodifyCLIP's emotional intelligence, this training-free system enables precise emotional transformations while preserving content integrity.

% This carefully designed framework enhances emotion-driven content creation by enabling intuitive, controllable, and high-fidelity image editing, 
%completes and enhances the entire emotional content creation pipeline, 
% as proven quantitatively and qualitatively in Section \ref{sec:experiments}.

%% file: sec/2_related_works.tex
\vspace{-2pt}
\section{Related Works}

\textbf{Visual Emotion Datasets.}
The discrete categorical system proposed by Ekman~\cite{ekman1987universals} identifies a set of universally recognized affective states, such as happiness, sadness, anger, and fear. These states are considered discrete categories that capture different emotional experiences.
Numerous datasets have been constructed to study people's emotional reactions to images~\cite{machajdik2010affective, peng2015mixed, yang2023emoset, kosti2019context, achlioptas2023affection}. However, these existing datasets often fail to capture complex or nuanced emotions, and their limited range of emotions can reduce their effectiveness in downstream tasks.
%such as sentiment analysis. 
%have limitations regarding the range of emotion dimensions they cover. Specifically:
%\textbf{1). Inaccuracy in Emotion Representation:} Existing datasets may fail to capture complex or nuanced emotions.
%\textbf{2). Limited Utility for Downstream Tasks:} The restricted range of emotions can hinder applications like sentiment analysis.
Therefore, there is a need for more comprehensive datasets that encompass a broader spectrum of emotional experiences beyond basic emotions.

\textbf{Contrastive Language-Image Pre-training (CLIP)}~\cite{radford2021learning} leverages contrastive learning between text-image pairs, demonstrating robust zero-shot capabilities across detection~\cite{guopen, li2022grounded}, segmentation~\cite{lilanguage, xu2022groupvit}, video understanding~\cite{Luo2021CLIP4ClipAE,Xu2021VideoCLIPCP}, and image synthesis~\cite{Crowson2022VQGANCLIPOD, frans2022clipdraw, Ramesh2022HierarchicalTI, vinker2022clipasso}. Despite its versatility, recent studies~\cite{Kim2023RegionAwarePF, zeng2022multi} highlight CLIP’s limitations in fine-grained feature extraction and these token-region alignment methods remain ineffective for complex, longer captions. Synthetic data has improved vision-language pretraining in numerous works~\cite{fan2023improving, tian2023stablerep, Yang2023ALIPAL, pmlr-v202-zhao23l}, leveraging text-to-image models and LLMs to generate high-quality image-text pairs that address issues like misalignment and harmful content in web-scraped datasets. While recent approaches~\cite{Lai2023VeCLIPIC, liu2023mllms} use multimodal LLMs for better caption generation, they mainly focus on short, descriptive captions rather than capturing the rich emotional and affective dimensions that longer, more nuanced captions could provide. 

\textbf{Image Editing and Attention Control.} Text-guided image editing has evolved from limited GAN frameworks~\cite{Goodfellow2014GenerativeAN, Reed2016GenerativeAT} to diffusion models~\cite{ho2020denoising, ramesh2022hierarchical, saharia2022photorealistic, rombach2022high} with cross-modal attention control~\cite{Meng2021SDEditGI, hertz2022prompt, Kawar2022ImagicTR, gu2023photoswap}. This progression has improved the accessibility of visual content manipulation, allowing users to transform images through simple textual descriptions rather than intricate manual editing. While traditional mask-based techniques~\cite{Nichol2021GLIDETP, Avrahami2021BlendedDF, Wang2022ImagenEA, bar2022text2live, couairon2023diffedit} require precise region selection, newer attention-based methods such as Prompt-to-Prompt~\cite{hertz2022prompt}, Plug-and-Play~\cite{Tumanyan2022PlugandPlayDF}, and MasaCtrl~\cite{Cao2023MasaCtrlTM} enable zero-shot editing by preserving structural integrity or conceptual elements without explicit boundary definition. These innovations strategically modify attention between text and visual features for coherent transformations. Though instruction-tuned models~\cite{Brooks2022InstructPix2PixLT, Geng23instructdiff} have advanced, they struggle with complex emotional transformations. Our approach leverages MLLMs for explicit emotional guidance and enhanced attention control.

%% file: sec/3_method.tex
% \section{Final copy}

% You must include your signed IEEE copyright release form when you submit your finished paper.
% We MUST have this form before your paper can be published in the proceedings.

% Please direct any questions to the production editor in charge of these proceedings at the IEEE Computer Society Press:
% \url{https://www.computer.org/about/contact}.

\vspace{-2pt}
\section{Methodology}
\vspace{-2pt}
\subsection{MoodArchive}
Language-image pre-training faces a key limitation: existing datasets lack detailed captions that effectively capture the complex emotional dimensions of visual content, hindering models’ ability to develop nuanced affective understanding. To address this, we introduce MoodArchive, a dataset with over 8 million images featuring diverse emotional content. The data collection steps are as follows: 

\textbf{\includegraphics[height=1.2em]{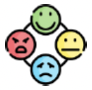} Comprehensive list of 27 emotions in 4 contexts. } 
As showcased in Fig. 1A in the supplementary material (Supp.), to ensure our dataset surpasses existing datasets ~\cite{machajdik2010affective, peng2015mixed, yang2023emoset, kosti2019context, achlioptas2023affection} in both scale and the granularity of emotion classes, we adopted the emotion list from GoEmotions~\cite{demszky2020goemotions}
(27 emotions\footnote{admiration, amusement, approval, caring, desire, excitement, gratitude, joy, love, optimism, pride, relief, anger, annoyance, disappointment, disapproval, disgust, embarrassment, fear, nervousness, grief, remorse, sadness, confusion, curiosity, realization, surprise}).
For each emotion, we aim to collect images in four distinct contexts: facial expressions, natural scenery, urban scenery, object classes. 

% \textbf{\includegraphics[height=1.2em]{figs/chatgpt.png} Coarse-to-fine-grained emotion-triggering context expansion with ChatGPT. } To effectively search for images, we sought assistance from ChatGPT to expand and concretize each emotion category. This approach was motivated by the difficulty of using abstract concepts like ``anger" to retrieve relevant images. Searching with a broad term such as "anger" often produces coarse and noisy results. To mitigate this, ChatGPT helped decompose high-level emotions into specific, real-life scenarios. For example, in Fig. 1A in Supp., ``anger" was further detailed into descriptors like 'pursed lips', 'glaring eyes', 'furrowed brows', and 'tightened jaw' under the facial expression subcategory, while examples such as 'fierce wildfire' and 'flood rampaging' were generated for natural scenery. For additional details on other emotions, please refer to the Supp. 
\textbf{\includegraphics[height=1.2em]{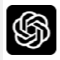} Emotion Context Expansion.} We used ChatGPT to decompose abstract emotions into specific descriptors, addressing the challenge of retrieving relevant images with broad terms like "anger." This yielded precise descriptors such as 'pursed lips' and 'glaring eyes' for facial expressions, and 'fierce wildfire' for natural scenes (See Fig. 1 in supp for additional details).
% \textbf{\includegraphics[height=1.2em]{figs/retrieval.png} Image Retrieval.} 
These refined phrases served as search queries to retrieve images from multiple sources\footnote{\href{https://www.unsplash.com/}{unsplash}, \href{https://www.pexels.com/}{pexels}, \href{https://www.pixabay.com/}{pixabay}, \href{https://elements.envato.com/}{elements.envato}, \href{https://openverse.org/}{openverse}, \href{https://burst.shopify.com/}{burst.shopify}, \href{https://stocksnap.io/}{stocksnap.io}, \href{https://www.foodiesfeed.com/}{foodiesfeed}, \href{https://freenaturestock.com/}{freenaturestock}}. 

% \textbf{\includegraphics[height=1.2em]{figs/retrieval.png} Image Retrieval.} The detailed scenarios and refined key phrases from ChatGPT were then used as precise search queries to retrieve 
% %With detailed scenarios and key phrases(or short sentences) from ChatGPT, we use those to retrieve 
% a large number of relevant images from varied sources \footnote{\href{https://www.unsplash.com/}{unsplash}, \href{https://www.pexels.com/}{pexels}, \href{https://www.pixabay.com/}{pixabay}, \href{https://elements.envato.com/}{elements.envato}, \href{https://openverse.org/}{openverse}, \href{https://burst.shopify.com/}{burst.shopify}, \href{https://stocksnap.io/}{stocksnap.io}, \href{https://www.foodiesfeed.com/}{foodiesfeed}, \href{https://freenaturestock.com/}{freenaturestock}}. 

\begin{figure}[tbp]
\centering
\includegraphics[width=\columnwidth]{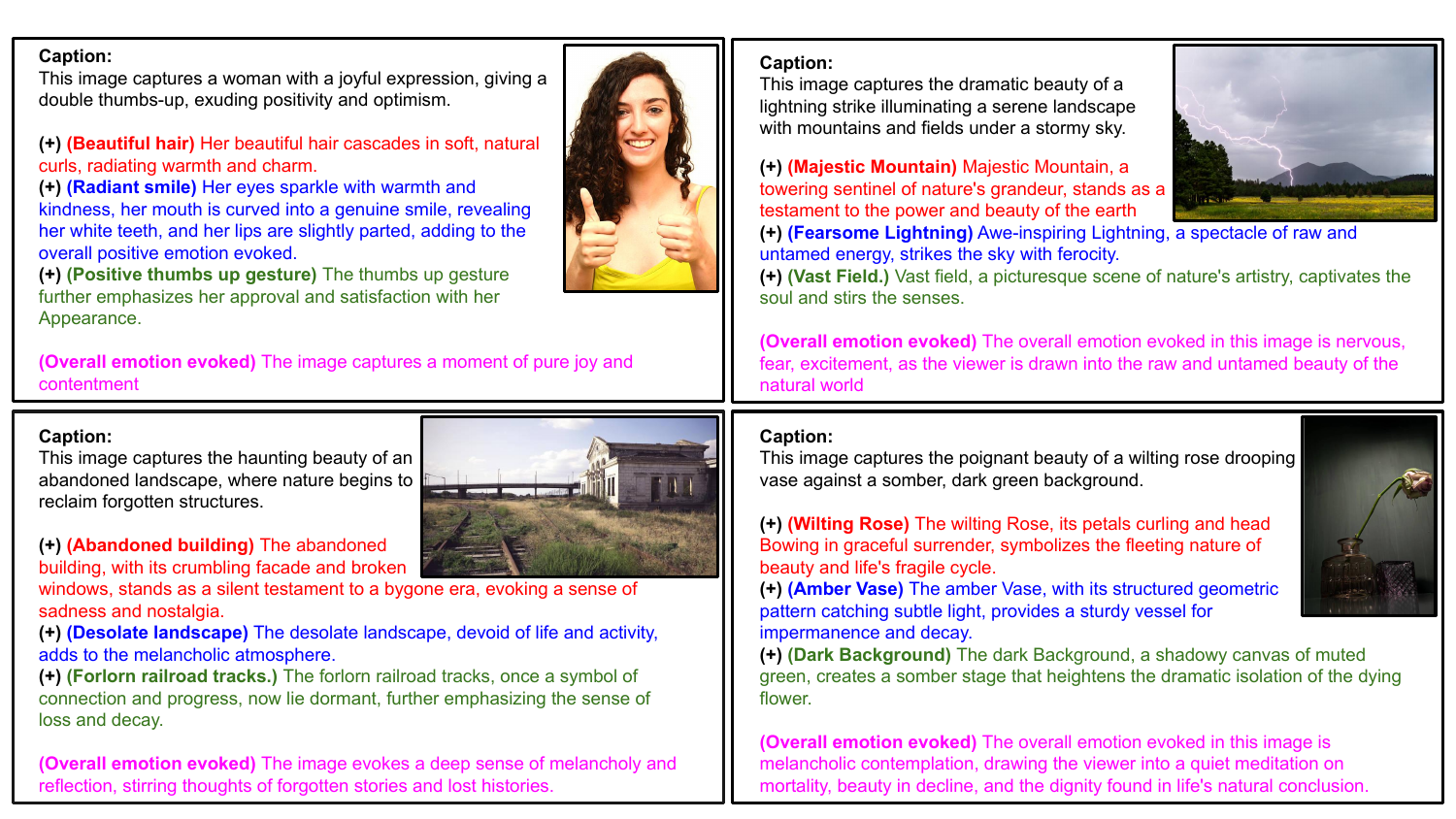}
\caption{Each image in MoodArchive is paired with a structured caption generated by LLaVA-NeXT, including (1) a concise global summary, (2) three local emotional stimuli (i.e. visual elements evoking specific emotions), and (3) an overall emotion assessment. (See Supp. for instruction prompt.)
%Each image in MoodArchived is paired with a caption generated by LLaVA-NeXT. The caption format for each image includes: 
%caption format we assigned to images collected in %MoodArchive. Our structured captioning includes: 
%(1) a concise global summary, (2) three local emotional stimuli elements (i.e. visual elements in the image that evoke specific emotions), and (3) an overall emotion assessment. (See Supp. for instruction prompt.)
% This structure balances semantic content with emotional impact, avoiding both over-simplification and irrelevant details.
}
\label{fig:moodifyclip_recap}
\vspace{-10pt}
\end{figure}

\textbf{\includegraphics[height=1.2em]{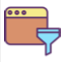} Post-processing.} We filtered NSFW content and corrupt files, then generated structured annotations using LLaVA-NeXT~\cite{liu2024visual}. Each image received: (1) a global summary, (2) three emotional stimuli identifying visual triggers, and (3) an overall emotion assessment (Fig. \ref{fig:moodifyclip_recap}). Finally, we filtered out the bottom 20\% of images based on CLIP score similarity to emotion-related phrases.

 % Finally, to refine the dataset further, we filtered out less relevant images for each emotion. Semantic similarity was assessed using CLIP scores between images and emotion-related key phrases, and the bottom 20\% of low-relevance images were excluded.
 
%Lastly, to further refine the cleaned set, we need to filter out those less relevant images for each desired emotion. The semantic similarity between them is assessed by the clip score between the images and the emotion-related key phrases generated earlier. Images that fell into the bottom 20\% of relevance scores were excluded.

\textbf{\includegraphics[height=1.2em]{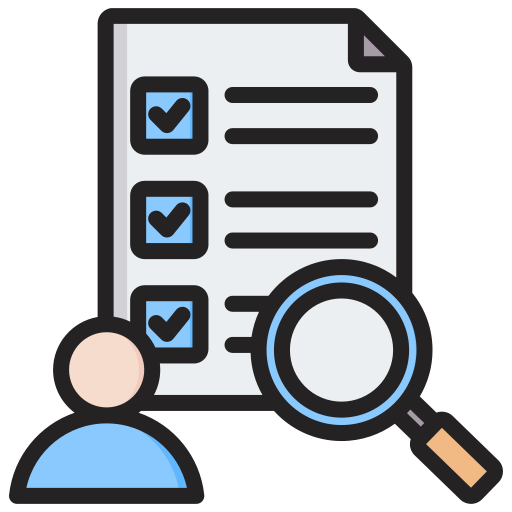} Human Validation Study.}
To evaluate the quality and reliability of our dataset, we conducted a comprehensive human validation study on Amazon Mechanical Turk. We randomly selected 10k images from MoodArchive and asked workers to compare the original web-collected alt-text with our LLaVA-generated detailed captions. The validation interface (Fig. 2 in Supp.) enforced strict evaluation criteria, requiring workers to assess both content accuracy and emotional interpretation. Captions were rejected if any component in the structured annotation (Fig. \ref{fig:moodifyclip_recap}) was inaccurate, ensuring a rigorous standard.
Results of the human evaluation showed that 85\% of the LLaVA-generated captions were selected by workers as better describing the images than the original web-collected alt-text, while the remaining 15\% were rejected in favor of either the original captions or indicated as inadequate, in which cases workers provided specific feedback citing issues such as inappropriate word choices, inaccurate emotion detection, overly dramatic descriptions, or misalignment between the detected emotions and cultural interpretations. This study demonstrates that our pipeline produces annotations that align well with human judgment while maintaining scalability. 
By integrating automated annotation generation with human verification, we create a large-scale emotional dataset that balances quality with scalability.
\subsection{MoodifyCLIP}

Although our human validation study confirms the decent quality of LLaVA-generated captions, we acknowledge that hallucinations are inevitable at this scale. 
However, 
% their rich emotional context provides invaluable nuance. To handle longer descriptions, we adopted positional embedding interpolation~\cite{Zhang2024LongCLIPUT} and balance accuracy with detail through: (1) concise summary captions offsetting potential inaccuracies, (2) fine-grained region-emotion alignment, and (3) optimal transport loss filtering mismatches between modalities.
the fine-grained emotional context these rich descriptions provide is invaluable, capturing nuances that shorter captions cannot. To handle longer descriptions, we adopt a positional embedding interpolation strategy~\cite{Zhang2024LongCLIPUT}. To balance accuracy and detail, MoodifyCLIP incorporates: (1) concise summary captions to offset potential inaccuracies in detailed descriptions, (2) fine-grained alignment to precisely link image regions with emotional annotations, and (3) optimal transport loss to filter mismatches between visual and textual features. Our experimental results confirm that the emotional understanding gained far outweighs occasional inaccuracies.

\textbf{Global Contrastive Loss (GL).} 
We adopt CLIP's visual encoder $E_i$ and text encoder $E_t$ to map inputs into a unified embedding space: $I^f = E_i(I), \quad T^f = E_t(T^f)$. 
% To balance detailed emotional triggers with global understanding, we incorporate both full complete captions $T^f$ and summary captions $T^s$ (first and last sentences only, see Fig. \ref{fig:moodifyclip_recap}). We compute the text-to-vision loss via InfoNCE contrastive learning: $\mathcal{L}_f = (\mathcal{L}_{t2v}^f + \mathcal{L}_{v2t}^f) / 2$
% Our MoodifyCLIP leverages the CLIP architecture, a visual encoder $E_i$ and a text encoder $E_t$ mapping inputs into a unified embedding space. We extract distinctive representations through the encoders:
% $I^f = E_i(I), \quad T^f = E_t(T)$.
While detailed captions capture specific emotional triggers and nuanced affective dimensions missing from existing datasets, concise summary captions remain crucial for holistic image comprehension. To balance granular details with global understanding, we add $T^s = E_t(T^s)$. Note that $T^f$ contains the complete five-sentence caption from our LLaVA annotations, while $T^s$ only includes the first and last sentences (i.e. summary caption and emotion assessment, see Fig. \ref{fig:moodifyclip_recap}).
% %Incorporating both description levels simultaneously grasp overall emotional tone and identify specific visual elements evoking those emotions. 
As outlined in Algorithm 1 in the Supp., we obtain the text-to-vision loss via InfoNCE contrastive learning between visual representations $I^f$ and textual embeddings $T^f$ as:
$\mathcal{L}_f = (\mathcal{L}_{t2v}^f + \mathcal{L}_{v2t}^f) / 2$
with explicit formulation :
\begin{equation}
\begin{aligned}
\mathcal{L}_{t2v}^f &= -\sum_{i=1}^N \log \frac{\exp(\cos(I_i^f, T_i^f)/\tau)}{\sum_{j=1}^N \exp(\cos(I_j^f, T_i^f)/\tau)} \\
\mathcal{L}_{v2t}^f &= -\sum_{i=1}^N \log \frac{\exp(\cos(T_i^f, I_i^f)/\tau)}{\sum_{j=1}^N \exp(\cos(T_j^f, I_i^f)/\tau)}
\end{aligned}
\end{equation}

Similarly, we add an objective for the global summary caption and emotion asessment:
%And we add a parallel objective for abbreviated captions:
$\mathcal{L}_s = (\mathcal{L}_{t2v}^s + \mathcal{L}_{v2t}^s) / 2$.

\textbf{Fine-grained Loss (FG).} Next, to fully utilize our hierarchical annotations, where each emotional stimulus prompt is carefully engineered to describe distinct affective elements, we align each with its corresponding regions of interest within the image.
We first compute cross-attention weights between fine-grained text embeddings ($T^{fg}$) and image patch embeddings ($I^{fg}$), yielding a similarity matrix $\mathcal{W} = \{w_{i,j}\}$. Applying these weights ensures that each emotional stimulus description is primarily focused on its relevant visual regions while differentiating itself from others. The attention-weighted visual representation for each stimulus $j$ is computed as $I_j^{fg} = \sum_{m=1}^{M} \frac{w_{i,j}}{\sum_j^3 w_{i,j}} I_m^{fg}$. 
We implement a contrastive objective for such region visual representations as:

% \vspace{-0.5em}

\begin{equation}
\begin{aligned}
\mathcal{L}^{fg}_{t2v} = -\sum_{i=1}^{N} \sum_{j=1}^{3} \log \frac{\exp(\cos(I^{fg}_{i,j}, T^{fg}_{i,j})/\tau)}{\sum_{m=1}^{M} \exp(\cos(I^{fg}_{i,m}, T^{fg}_{i,j})/\tau)}\\
\mathcal{L}^{fg}_{v2t} = -\sum_{i=1}^{N} \sum_{j=1}^{3} \log \frac{\exp(\cos(T^{fg}_{i,j}, I^{fg}_{i,j})/\tau)}{\sum_{m=1}^{M} \exp(T^{fg}_{i,j}, \cos(I^{fg}_{i,m})/\tau)}
\end{aligned}
\end{equation}

Symmetrically we get $\mathcal{L}_{fg} = (\mathcal{L}_{t2v}^{fg} + \mathcal{L}_{v2t}^{fg}) / 2$.
This fine-grained alignment mechanism helps develop a sophisticated understanding of which visual elements trigger specific emotional responses (see implementation details in Algorithm 1 and Fig. 2 in Supp.).

\textbf{Optimal Transport Loss (OT).} Now that we have the fine-grained loss for specific connections (zoom-in), such as matching a smiling face region with text about happiness or tearful eyes with sadness, we take a bigger-picture (zoom-out) via optimal transport~\cite{shi2024ot}. Mathematically, it solves the matching problem by finding the most efficient overall assignment between all image regions and all text descriptions while respecting their similarity as in the matrix $\mathbf{S}= \langle \mathbf{I}_{\text{fg}}, \mathbf{T}_{\text{fg}} \rangle_{\text{batch}}$, which translates into transportation costs (distance matrix) as: $\mathbf{W} = \mathbf{1} - \mathbf{S}$.
Then, by asking the question, \textit{"What's the most efficient way to match all the emotional elements in these images with all the emotional concepts in these texts?"}, we use the Sinkhorn algorithm to find the optimal global assignment while respecting local emotional nuances. 
\begin{align}
\mathbf{T}_{ot} &= \text{sinkhorn}(\mathbf{W}) \\
&= \arg\min_{\mathbf{P} \in \Pi(\mathbf{a},\mathbf{b})} \langle \mathbf{W}, \mathbf{P} \rangle - \epsilon H(\mathbf{P})
\end{align}
where $\mathbf{a}$
and $\mathbf{b}$ are uniform distributions over image regions and text features respectively, while $\Pi(\mathbf{a},\mathbf{b})$ is the set of all possible transport plans with these marginals, and $H(\mathbf{P})$ is an entropy regularization term.

Here higher values in $T$ represent preferred transport paths (where it's ``cheaper" to move mass), hence amplifying signals from high-similarity pairs while downplaying mismatches. This is particularly valuable for emotions, which are often ambiguous and overlapping.
The final OT loss is computed as $\mathcal{L}_{ot} = \text{CrossEntropy}(\mathbf{T}_{\text{ot}} \odot \mathbf{S}, \mathbf{I})$,
where $\mathbf{I}$
 is the identity matrix (see implementation details in Algorithm 1 in the Supp.).

Our total loss combines all components as: 
\begin{equation}
\mathcal{L}_{\text{moodifyclip}}=\lambda_{f}\cdot\mathcal{L}_{f} + \lambda_{s}\cdot\mathcal{L}_{s} + \lambda_{fg}\cdot\mathcal{L}_{fg} + \lambda_{ot}\cdot\mathcal{L}_{ot}
\end{equation}

\subsection{Moodifier}

\begin{figure*}[htbp]
\centering
\includegraphics[width=\textwidth]{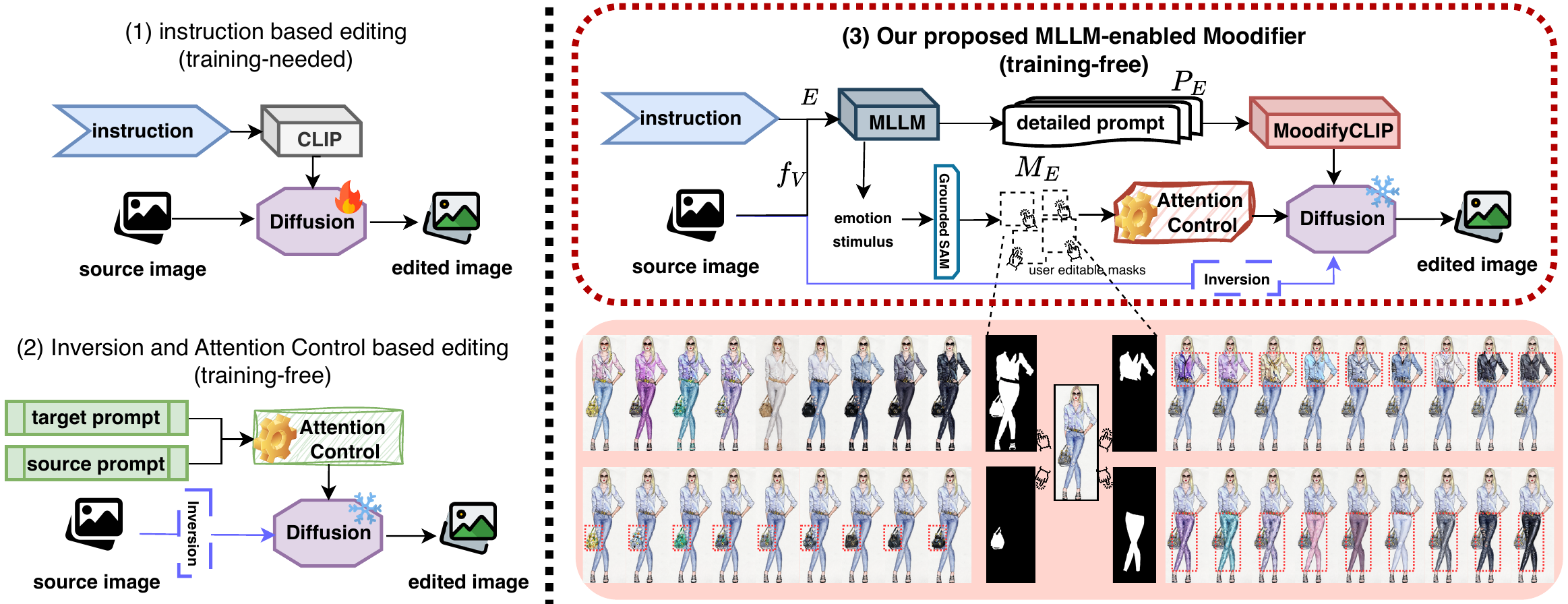}
\caption{
% Comparison of image editing approaches with our MLLM-enhanced Moodifier. Our MLLM-enhanced Moodifier combines ideas from training-required instruction-based methods and training-free attention control approaches, creating a system that automatically converts simple instructions into rich emotional prompts, then applies targeted edits using our specialized MoodifyCLIP and region-specific processing. The bottom panel demonstrates practical application, showing the fully automated pipeline while also offering flexibility for users to optionally select specific emotion-focused regions if desired. Unlike inpainting, which often introduces discontinuities, our method modifies the source image's latent representation to ensure natural transitions between edited and preserved regions.
Our MLLM-enhanced Moodifier (right panel (3)) integrates ideas from instruction-based and attention control approaches, converting simple commands into rich emotional prompts for targeted editing. We applied LLaVA-NeXT for the MLLM. The bottom of panel (3) demonstrates that our system offers both fully automated processing and optional region selection, while ensuring natural transitions by modifying latent representations instead of using inpainting. The left panels (1) and (2) summarize existing paradigms from previous works for comparison. 
}
\label{fig:moodifer_workflow}
\end{figure*}

With MoodifyCLIP fine-tuned on MoodArchive, we propose Moodifier, a training-free, emotion-driven image editing system. The Moodifier workflow is illustrated in Fig. \ref{fig:moodifer_workflow} and Algorithm \ref{algo:moodifier}. It first extracts visual features $f_V = \text{Enc}_{\text{vis}}(V)$ from the source image, enabling MLLM (in our case we applied LLaVA-NeXT) to generate emotion-specific prompts $P_E = \text{MLLM}(f_V, E)$ and attention maps $M_E$ that identify where modifications should occur (see Supp. for instruction prompts). These outputs then guide the diffusion process to produce the emotionally transformed image. More specificially, we adopt the non-iterative inversion approach from~\cite{xu2023inversion} to obtain the latent representation $z$.
Inspired by~\cite{hertz2022prompt}, we then manipulate the cross-attention mechanisms within the diffusion process to achieve precise emotional transformations. We operate on the key insight that cross-attention maps define the relationship between spatial image features and textual concepts.

\begin{algorithm}[h]
\caption{Moodifier: MLLM-Enhanced Emotional Editing}
\label{algo:moodifier}
\begin{algorithmic}[1]
\REQUIRE Source image $V$, target emotion $E$
\ENSURE Emotionally edited image $V'$
\STATE $f_V \leftarrow \text{Enc}_{\text{vis}}(V)$ 
% \COMMENT{Extract visual features}
\STATE $P_E, M_E \leftarrow \text{MLLM}(f_V, E))$ 
% \COMMENT{Generate guidance}
\STATE $P_E \leftarrow \text{MoodifyCLIP}(P_E)$
% \COMMENT{Interpret emotion}
\STATE $z \leftarrow \text{Inv}_{\text{dm}}(V)$
% \COMMENT{Inversion}
\STATE Initialize $z_T^* \leftarrow z$
% \COMMENT{Setup source latent}
\FOR{$t = T, T-1, \ldots, 1$} 
% \COMMENT{Diffusion process}
    \STATE $z_{t-1}, M_t \leftarrow \text{DM}(z_t, \emptyset, t)$
    % \COMMENT{Source attention maps}
    \STATE $M_t^* \leftarrow \text{DM}(z_t^*, P_E, t)$ 
    % \COMMENT{Target attention maps}
    \STATE $\tilde{M}_t \leftarrow \text{BlendMaps}(M_t, M_t^*, M_E, t)$ 
    % \COMMENT{Emotion-guided blend}
    \STATE $z_{t-1}^* \leftarrow \text{DM}(z_t^*, P_E, t)\{M \leftarrow \tilde{M}_t\}$ 
    % \COMMENT{Apply attention}
    % \IF{$\text{local\_control}$}
    \STATE $z_{t-1}^* \leftarrow 
    (1 - \mathbbm{1}_{M_E > 0}) \odot z_{t-1} + \mathbbm{1}_{M_E > 0} \odot z_{t-1}^*$
    % \text{LocalEdit}(z_{t-1}, z_{t-1}^*, M_E)$ 
    %     % \COMMENT{Local editing}
    % \ENDIF
\ENDFOR
\RETURN $z_0^*$ 
% \COMMENT{Return edited image}
\end{algorithmic}
\end{algorithm}

In text-conditioned diffusion models, each diffusion step computes attention maps $M_t$ through:
$M_t = \text{Softmax}\left(\frac{QK^T}{\sqrt{d}}\right)$,
where $Q = \ell_Q(\phi(z_t))$ represents visual feature queries, and $K = \ell_K(\psi(P))$ represents textual feature keys. The cell $M_{i,j}$ defines the influence of the $j$-th textual token on the $i$-th pixel.
For emotional editing, we need precise control over which image regions should change and how. Our MLLM generates not only detailed prompts $P_E$ but also spatial attention maps $M_E$ that identify emotion-relevant regions. As shown in Algorithm  \ref{algo:moodifier}, DM (Diffusion Model) refers to a single step of the diffusion process that denoises the latent representation while computing cross-attention maps between visual and textual features. Specifically, $\text{DM}(z_t, P_E, t)$ computes one denoising step from noise level $t$ to $t-1$ conditioned on prompt $P$, producing both the denoised latent $z_{t-1}$ and the attention maps $M_t$.
We blend attention maps from the source image with those generated for the target emotion:
$\tilde{M}_t = \text{BlendMaps}(M_t, M_t^*, M_E, t)$.
This blending function applies attention control to preserve structural information from the source image while enabling targeted emotional modifications guided by the auto-generated attention maps $M_E$:
\begin{equation}
    \tilde{M}_t = \begin{cases} \text{Refine}(M^{\text{src}}, M^{\text{tgt}}, M_E) & t \geq \tau_c \\ M^{\text{tgt}} & t < \tau_c \end{cases}
\end{equation}
\begin{equation}
    \text{Refine}(M^{\text{src}}, M^{\text{tgt}}, M_E)_{i,j} = \begin{cases} (M^{\text{tgt}}){i,j} & \text{if } (i,j) \in M_E \\ (M^{\text{src}})_{i,j} & \text{otherwise} \end{cases}
\end{equation}
where $\tau_c$ controls attention strength. 
At early steps ($t \geq \tau_c$), refined attention is used to establish structure while preserving identity, whereas in later steps ($t < \tau_c$), target attention maps enhance emotional details. 
% \textcolor{blue}{[WRAP UP NEEDED]}
Finally, the emotion stimulus maps $M_E$ smoothly blend source and target latents only in regions that should express the target emotion, while preserving the rest of the image.

%% file: sec/4_experiments.tex
\section{Experiments}\label{sec:experiments}

\subsection{Experiment Settings}

We evaluate our approach from two perspectives: (1) the zero-shot representation learning performance of the MoodifyCLIP model, and (2) the image editing performance of the Moodifier framework.

% \footnotetext[1]{$^\dagger$ CLIP$_{CommonPool}$: CLIP-ViT-B-16-CommonPool.L-s1B-b8K.}
% \footnotetext[2]{$^\ddagger$ CLIP$_{LAION}$: CLIP-ViT-B-16-laion2B-s34B-b88K.}
% \footnotetext[3]{$^\S$ CLIP$_{DataComp}$: CLIP-ViT-B-16-DataComp.XL-s13B-b90K.}
% \footnotetext[4]{$^\P$ CLIP$_{CommonPool}$: CLIP-ViT-L-14-CommonPool.XL-s13B-b90K.}
% \footnotetext[5]{$^{**}$ CLIP$_{LAION}$: CLIP-ViT-L-14-laion2B-s32B-b82K.}
% \footnotetext[6]{$^{\dagger\dagger}$ CLIP$_{DataComp}$: CLIP-ViT-L-14-DataComp.XL-s13B-b90K.}

% \begin{figure}[!htbp]
% \centering
% \includegraphics[width=0.47\textwidth]{figs/radar_and_bar_chart.png}
% \caption{Performance comparison of emotion editing methods. \textbf{Left:} Radar chart showing CLIP scores across object categories, with Moodifier (green) achieving the highest average score. \textbf{Right:} Face-specific metrics comparing PSNR and LPIPS ($\times10^2$) values, with Moodifier demonstrating optimal balance between semantic transformation and identity preservation.}
% \label{fig:category_clip_comparison}
% \end{figure}

\textbf{MoodifyCLIP Evaluation.}
We evaluate MoodifyCLIP on zero-shot emotion classification and image-text retrieval using 5-fold cross-validation. For classification, we test on Emotion6~\cite{peng2015mixed}, EmoSet~\cite{yang2023emoset}, and Emotic~\cite{kosti2019context}, reporting Top-1/2/3 accuracy against three CLIP variants in both B/16 and L/14 sizes. For retrieval, we evaluate on SentiCap~\cite{mathews2016senticap}, Affection~\cite{Achlioptas2022AffectionLA}, and MoodArchive-5k (5k human-verified pairs held out from training), reporting R@1 and R@5 for both retrieval directions.
% We evaluate MoodifyCLIP on two downstream tasks: zero-shot emotion classification and zero-shot image-text retrieval. For both tasks, we conduct 5-fold cross-validation and report mean performance with standard deviation to ensure robust evaluation.
% For classification, we test on three emotional datasets: Emotion6~\cite{peng2015mixed}, EmoSet~\cite{yang2023emoset}, and Emotic~\cite{kosti2019context}, reporting Top-1/2/3 accuracy (\%), which measures the percentage of images where the ground truth emotion appears among the model's top predictions. We compare our MoodifyCLIP against three CLIP variants (CLIP$_{CommonPool}$, CLIP$_{DataComp}$, CLIP$_{LAION}$) using both B/16 and L/14 model sizes.
% For image-text retrieval, we evaluate on SentiCap~\cite{mathews2016senticap}, Affection~\cite{Achlioptas2022AffectionLA}, and MoodArchive-5k (a curated test set of 5k image-text pairs randomly selected from our human-verified subset of MoodArchive and held out from training).
% %It follows the same hierarchical captioning format illustrated in Fig. \ref{fig:moodifyclip_recap}, which explains its relatively higher performance across all models. 
% We report Recall@1 (R@1) and Recall@5 (R@5) for both Image-to-Text and Text-to-Image retrieval, measuring the percentage of queries where the correct item appears in the top K retrieved results.

\textbf{Moodifier Evaluation.}
Evaluating emotion-driven image editing presents a fundamental challenge: the absence of standardized ground truth. Unlike tasks with well-defined objectives (e.g., “add a cat”), emotional transformations lack clear reference images. Traditional fidelity metrics such as PSNR and SSIM remain essential for ensuring that edited images preserve content integrity, preventing cases where emotional accuracy is achieved at the cost of extreme distortion or identity loss. 

To balance structure-preserving evaluations with emotion-specific assessments, we employ a two-pronged approach for evaluating Moodifier: (1) quantitative metrics measuring both preservation and transformation aspects, and (2) human evaluation to assess visual appeal and emotional accuracy.
% We conduct a comprehensive evaluation of our Moodifier framework using a diverse test set of 240 images (30 images per category across 8 categories: face, bag, bracelet, clothes, earring, necklace, ring, and vase) from MoodArchive, ensuring no overlap with the training data for MoodifyCLIP. For each test (or source) image, we generate 27 distinct emotion-transformed versions corresponding to our emotion taxonomy (detailed in Fig. 1 in Supp.). 
Our metrics are computed hierarchically: using 240 images (30 images per category across 8 categories: face, bag, bracelet, clothes, earring, necklace, ring, and vase from MoodArchive, no overlap with MoodifyCLIP training set) as source, first at the source-synthesized pair level for each emotion, then averaged across all 27 emotions per source image, then across all 30 images per category, and finally across all 8 categories to obtain overall model-level performance. 
We measure structural distance between the source and synthesized images' structural features; background preservation through PSNR, LPIPS, MSE, and SSIM; emotional transformation via the CLIP Similarity score between synthesized images and target emotions; and human evaluation through MTurk assessments (see more details in Fig. 4 and Sec. 5.3 in Supp.). While CLIP may overlook subtle nuances and human evaluation is inherently subjective, their combination provides complementary insights, ensuring both accuracy and practical usability. As shown in Table \ref{tab:comparison}, we compare our Moodifier against several state-of-the-art image editing methods~\cite{Zhang2023AddingCC, Brooks2022InstructPix2PixLT, Geng23instructdiff, Cao2023MasaCtrlTM, Tumanyan2022PlugandPlayDF, HubermanSpiegelglas2023AnEF, Wang2023StyleDiffusionCD, 
mokady2023null, 
ju2024pnp,
xu2023inversion}.

\subsection{Result Analysis}

\begin{table*}[!htbp]
\centering
\resizebox{\textwidth}{!}{
\begin{tabular}{c|l|c c c|c c c|c c c}
\hline
& & \multicolumn{3}{c|}{\textbf{Emotion6}} & \multicolumn{3}{c|}{\textbf{EmoSet}} & \multicolumn{3}{c}{\textbf{Emotic}} \\
& & Top-1 & Top-2 & Top-3 & Top-1 & Top-2 & Top-3 & Top-1 & Top-2 & Top-3 \\
\hline
\multirow{4}{*}{B/16} 
& CLIP$_{CommonPool}$ & 41.82\textsubscript{±1.72} & 68.99\textsubscript{±1.80} & 83.13\textsubscript{±0.79} & 40.07\textsubscript{±0.24} & 62.73\textsubscript{±0.26} & 74.43\textsubscript{±0.34} & 27.59\textsubscript{±0.44} & 39.67\textsubscript{±0.75} & 48.74\textsubscript{±0.57} \\
& CLIP$_{DataComp}$ & 47.98\textsubscript{±1.53} & 71.92\textsubscript{±1.61} & 86.57\textsubscript{±1.92} & 45.88\textsubscript{±0.35} & 65.33\textsubscript{±0.32} & 76.73\textsubscript{±0.25} & 24.28\textsubscript{±0.87} & 36.98\textsubscript{±0.44} & 46.21\textsubscript{±0.43} \\
& CLIP$_{LAION}$ & 50.30\textsubscript{±1.05} & 73.23\textsubscript{±3.00} & 84.80\textsubscript{±3.11} & \textbf{49.43}\textsubscript{±0.43} & \textbf{70.74}\textsubscript{±0.40} & \textbf{81.96}\textsubscript{±0.20} & 23.94\textsubscript{±0.61} & 35.70\textsubscript{±0.57} & 44.01\textsubscript{±0.39} \\
& MoodifyCLIP (Ours) & \textbf{55.86}\textsubscript{±0.84} & \textbf{77.93}\textsubscript{±1.72} & \textbf{90.66}\textsubscript{±1.50} & 46.96\textsubscript{±1.31} & 69.58\textsubscript{±1.93} & 81.22\textsubscript{±1.32} & \textbf{38.35}\textsubscript{±0.76} & \textbf{53.49}\textsubscript{±0.84} & \textbf{61.95}\textsubscript{±0.56} \\
\hline
\multirow{4}{*}{L/14} 
& CLIP$_{CommonPool}$ & 49.95\textsubscript{±0.92} & 74.75\textsubscript{±2.31} & 87.02\textsubscript{±1.73} & 46.62\textsubscript{±0.25} & 68.71\textsubscript{±0.31} & 80.49\textsubscript{±0.31} & 27.55\textsubscript{±0.49} & 41.55\textsubscript{±0.52} & 50.34\textsubscript{±0.64} \\
& CLIP$_{DataComp}$ & 51.77\textsubscript{±2.48} & 73.94\textsubscript{±1.52} & 87.63\textsubscript{±1.13} & 48.54\textsubscript{±0.38} & 67.43\textsubscript{±0.28} & 78.63\textsubscript{±0.27} & 26.02\textsubscript{±0.75} & 39.20\textsubscript{±0.83} & 47.95\textsubscript{±0.67} \\
& CLIP$_{LAION}$ & \textbf{55.76}\textsubscript{±0.29} & 79.39\textsubscript{±2.27} & 89.80\textsubscript{±1.72} & 49.98\textsubscript{±0.40} & 70.60\textsubscript{±0.40} & 81.83\textsubscript{±0.30} & 21.53\textsubscript{±0.26} & 33.07\textsubscript{±0.20} & 41.53\textsubscript{±0.32} \\
& MoodifyCLIP (Ours) & 54.83\textsubscript{±2.32} & \textbf{80.51}\textsubscript{±0.98} & \textbf{91.11}\textsubscript{±0.79} & \textbf{56.33}\textsubscript{±0.29} & \textbf{72.49}\textsubscript{±0.27} & \textbf{83.56}\textsubscript{±0.26} & \textbf{38.40}\textsubscript{±0.79} & \textbf{57.09}\textsubscript{±0.78} & \textbf{67.12}\textsubscript{±0.82} \\
\hline
\end{tabular}
}
\caption{Results of MoodifyCLIP for zero-shot classification across three emotional datasets (Emotion6, EmoSet, and Emotic). Top-1, Top-2, and Top-3 accuracy (\%) are reported. Bold numbers indicate the best performance. Details of specific CLIP variants in Supp.}
\label{tab:zero-shot-classification}
\end{table*}

\begin{table*}[!htbp]
\resizebox{\textwidth}{!}{
\begin{tabular}{c|l|c c|c c||c c|c c||c c|c c}
\hline
& & \multicolumn{4}{c||}{\textbf{SentiCap}} & \multicolumn{4}{c||}{\textbf{Affection}} & \multicolumn{4}{c}{\textbf{MoodArchive-5k}} \\
& & \multicolumn{2}{c|}{Image-to-Text} & \multicolumn{2}{c||}{Text-to-Image} & \multicolumn{2}{c|}{Image-to-Text} & \multicolumn{2}{c||}{Text-to-Image} & \multicolumn{2}{c|}{Image-to-Text} & \multicolumn{2}{c}{Text-to-Image} \\
& & R@1 & R@5 & R@1 & R@5 & R@1 & R@5 & R@1 & R@5 & R@1 & R@5 & R@1 & R@5 \\
\hline
\multirow{4}{*}{B/16} 
& CLIP$_{CommonPool}$ & 29.98\textsubscript{±0.98} & 65.08\textsubscript{±1.61} & 28.08\textsubscript{±1.02} & 63.68\textsubscript{±1.96} & 26.70\textsubscript{±0.41} & 54.02\textsubscript{±0.83} & 25.36\textsubscript{±1.09} & 52.18\textsubscript{±0.55} & 79.22\textsubscript{±1.02} & 94.56\textsubscript{±0.73} & 69.60\textsubscript{±0.89} & 89.26\textsubscript{±0.92} \\
& CLIP$_{DataComp}$ & 41.08\textsubscript{±1.45} & 78.46\textsubscript{±0.93} & 39.74\textsubscript{±0.81} & 76.22\textsubscript{±0.58} & 40.30\textsubscript{±1.55} & 66.98\textsubscript{±0.50} & 37.74\textsubscript{±0.92} & 63.96\textsubscript{±1.45} & 82.90\textsubscript{±0.98} & 98.42\textsubscript{±0.25} & 77.44\textsubscript{±0.92} & 96.18\textsubscript{±0.57} \\
& CLIP$_{LAION}$ & \textbf{42.54\textsubscript{±0.57}} & \textbf{79.24\textsubscript{±1.03}} & \textbf{41.24\textsubscript{±0.36}} & 77.80\textsubscript{±1.05} & 40.78\textsubscript{±0.69} & 68.20\textsubscript{±1.16} & 39.66\textsubscript{±0.56} & 66.84\textsubscript{±1.17} & 83.22\textsubscript{±0.69} & 98.24\textsubscript{±0.34} & 78.40\textsubscript{±1.14} & 96.62\textsubscript{±0.44} \\
& MoodifyCLIP (Ours) & 41.10\textsubscript{±1.30} & 78.46\textsubscript{±0.50} & 40.40\textsubscript{±0.81} & \textbf{78.54\textsubscript{±1.37}} & \textbf{42.58\textsubscript{±1.41}} & \textbf{70.60\textsubscript{±1.20}} & \textbf{43.70\textsubscript{±0.73}} & \textbf{72.58\textsubscript{±1.05}} & \textbf{84.52\textsubscript{±0.46}} & \textbf{99.02\textsubscript{±0.51}} & \textbf{81.20\textsubscript{±0.71} }& \textbf{97.96\textsubscript{±0.45}} \\
\hline
\multirow{4}{*}{L/14} 
& CLIP$_{CommonPool}$ & 41.94\textsubscript{±0.33} & 78.54\textsubscript{±0.59} & 41.08\textsubscript{±1.13} & 77.40\textsubscript{±0.62} & 41.64\textsubscript{±1.23} & 69.16\textsubscript{±1.05} & 40.94\textsubscript{±0.72} & 67.44\textsubscript{±1.30} & 83.46\textsubscript{±1.08} & 98.48\textsubscript{±0.32} & 77.96\textsubscript{±1.21} & 95.94\textsubscript{±0.50} \\
& CLIP$_{DataComp}$ & 43.72\textsubscript{±1.04} & 80.08\textsubscript{±0.97} & 42.46\textsubscript{±0.30} & 79.18\textsubscript{±0.47} & 44.14\textsubscript{±0.87} & 70.96\textsubscript{±1.32} & 42.84\textsubscript{±1.77} & 69.10\textsubscript{±1.04} & 84.04\textsubscript{±0.96} & 98.62\textsubscript{±0.50} & 79.92\textsubscript{±0.82} & 96.84\textsubscript{±0.45} \\
& CLIP$_{LAION}$ & 43.94\textsubscript{±0.52} & 80.38\textsubscript{±1.22} & 42.40\textsubscript{±0.89} & 79.62\textsubscript{±1.46} & 43.90\textsubscript{±1.44} & 70.34\textsubscript{±1.24} & 44.44\textsubscript{±0.29} & 69.66\textsubscript{±1.74} & 85.02\textsubscript{±0.77} & 98.68\textsubscript{±0.29} & 81.28\textsubscript{±0.15} & 97.20\textsubscript{±0.49} \\
& MoodifyCLIP (Ours) & \textbf{44.72\textsubscript{±1.10}} & \textbf{80.46\textsubscript{±1.14}} & \textbf{44.90\textsubscript{±0.86}} & \textbf{82.18\textsubscript{±0.70}} & \textbf{46.34\textsubscript{±0.66}} & \textbf{74.02\textsubscript{±0.82}} & \textbf{48.56\textsubscript{±0.91}} & \textbf{76.56\textsubscript{±1.22}} & \textbf{86.16\textsubscript{±0.37}} & \textbf{98.84\textsubscript{±0.38}} & \textbf{85.18\textsubscript{±0.54}} & \textbf{98.96\textsubscript{±0.29}} \\
\hline
\end{tabular}
}
\caption{Results of MoodifyCLIP for zero-shot image-text retrieval across three datasets (SentiCap, Affection, and MoodArchive-5k) with both Image-to-Text and Text-to-Image settings. Details of specific CLIP variants in Supp.}
\label{tab:results}
\end{table*}

\textbf{MoodifyCLIP Performance.} 
Tables \ref{tab:zero-shot-classification} and \ref{tab:results} show MoodifyCLIP consistently outperforms baseline CLIP models across emotion classification and retrieval tasks. Using 5-fold cross-validation, we observe significant improvements, especially on challenging datasets like Emotic. While all models perform better on MoodArchive-5k due to caption format similarity, improvements on SentiCap and Affection confirm generalization capability. Table \ref{tab:moodifyclip_ablation} shows both loss components contribute: fine-grained alignment identifies emotion-specific elements, while optimal transport optimizes global emotional matching, with their combination delivering strongest results.
% Tables \ref{tab:zero-shot-classification} and  \ref{tab:results} demonstrate that our MoodifyCLIP model consistently outperforms baseline CLIP models across both emotion classification and image-text retrieval tasks. Using 5-fold cross-validation, we observe statistically significant improvements across all datasets. For zero-shot classification, our model shows substantial advantages on challenging datasets such as Emotic. For image-text retrieval, while all models perform better on MoodArchive-5k due to caption format similarity with our training data, MoodifyCLIP's improvements on SentiCap and Affection datasets confirm its generalization capability. Table \ref{tab:moodifyclip_ablation} presents ablation study results evaluating the contribution of both loss components in MoodifyCLIP. Fine-grained alignment alone improves performance across all datasets by identifying emotion-specific visual elements, while Optimal Transport alone yields consistent improvements by optimizing global emotional matching. Combining both components delivers the strongest results across both tasks.

\begin{table*}[!htbp]
\centering
\resizebox{\textwidth}{!}{
\begin{tabular}{cc|ccc|ccc|ccc}
\cline{1-11}
\multicolumn{2}{c|}{} & \multicolumn{3}{c|}{Text Retrieval} & \multicolumn{3}{c|}{Image Retrieval} & \multicolumn{3}{c}{Classification} \\
\multicolumn{2}{c|}{CLIP Losses} & SentiCap & Affection & MoodArchive-5k & SentiCap & Affection & MoodArchive-5k & Emotion6 & EmoSet & Emotic \\
FG & OT & R@1 & R@1 & R@1 & R@1 & R@1 & R@1 & Acc.(\%) & Acc.(\%) & Acc.(\%) \\
\cline{1-11}
$\times$ & $\times$ & 41.14\textsubscript{±0.84} & 40.68\textsubscript{±1.30} & 83.56\textsubscript{±1.13} & 40.38\textsubscript{±1.20} & 42.76\textsubscript{±1.59} & 84.08\textsubscript{±0.48} & 48.03\textsubscript{±2.32} & 52.56\textsubscript{±0.13} & 34.97\textsubscript{±0.31} \\
\checkmark & $\times$ & 42.46\textsubscript{±0.89} & 42.64\textsubscript{±0.96} & 84.36\textsubscript{±0.90} & 40.46\textsubscript{±1.40} & 44.00\textsubscript{±0.75} & 84.58\textsubscript{±0.52} & 51.82\textsubscript{±1.40} & 54.10\textsubscript{±0.28} & 37.85\textsubscript{±0.62} \\
$\times$ & \checkmark & 41.56\textsubscript{±0.59} & 41.28\textsubscript{±0.71} & 84.20\textsubscript{±0.99} & 39.64\textsubscript{±0.97} & 42.16\textsubscript{±1.04} & 84.30\textsubscript{±0.55} & 48.78\textsubscript{±1.45} & 52.93\textsubscript{±0.22} & 35.50\textsubscript{±0.76} \\
\checkmark & \checkmark & \textbf{44.72}\textsubscript{±1.10} & \textbf{46.34}\textsubscript{±0.66} & \textbf{86.16}\textsubscript{±0.37} & \textbf{44.90}\textsubscript{±0.86} & \textbf{48.56}\textsubscript{±0.91} & \textbf{85.18}\textsubscript{±0.54} & \textbf{54.83}\textsubscript{±2.32} & \textbf{56.33}\textsubscript{±0.29} & \textbf{38.40}\textsubscript{±0.79} \\
\cline{1-11}
\end{tabular}
}
\caption{Ablation study of different CLIP losses added to MoodifyCLIP. We use ViT-L/14 as the image backbone.}
\label{tab:moodifyclip_ablation}
\end{table*}

\textbf{Moodifier Image Editing Performance.} 
Moodifier’s two-pronged evaluation, combining quantitative metrics and human assessment, highlights its strength in achieving high emotional accuracy (as evidenced by CLIP scores and human preference) while preserving structural integrity. This balance is essential for practical applications, ensuring accurate emotional transformations without compromising content recognition.

\begin{table*} [!htbp]
\centering 
\resizebox{\textwidth}{!}{ 
\begin{tabular}{c|c|c|c|c|c|c|c|c|c} 
\hline 
Method & Structure & \multicolumn{4}{c|}{Background Preservation} & \multicolumn{1}{c|}{CLIP Score$^*$} & \multicolumn{3}{c}{Human Preference (\%)} \\ 
\hline  
& Distance$_{\times 10^3}$ $\downarrow$ & PSNR $\uparrow$ & LPIPS$_{\times 10^3}$ $\downarrow$ & MSE$_{\times 10^4}$ $\downarrow$ & SSIM$_{\times 10^2}$ $\uparrow$ & Similarity $\uparrow$ & Visual Appeal $\uparrow$ & Emotion Acc $\uparrow$& Overall $\uparrow$\\ 
\hline 
ControlNet & 21.9 & 22.01 & 102.3 & 104.8 & 80.3 & 12.80 & 28.9 & 24.7 & 26.8 \\ 
\hline 
Instruct-Pix2Pix & 22.0 & 23.17 & 112.6 & 100.2 & 81.0 & 13.27 & 38.5 & 35.8 & 34.3 \\ 
Instruct-Diffusion & 34.5 & 18.55 & 175.8 & 227.6 & 74.2 & 13.24 & 19.3 & 20.5 & 19.8 \\ 
\hline 
MasaCtrl & 18.4 & 21.52 & 144.2 & 76.2 & 79.4 & 12.80 & 27.4 & 26.3 & 24.9 \\ 
Plug-and-Play & \textbf{13.9} & 22.02 & 123.5 & \textbf{72.3} & 80.5 & 12.85 & 36.2 & 33.5 & 34.8 \\ 
Edit Friendly & 15.1 & 21.39 & 130.7 & 104.9 & 78.9 & 13.47 & 28.6 & 26.9 & 27.5 \\ 
StyleDiffusion & 20.4 & 17.42 & 159.7 & 219.0 & 70.9 & 13.42 & 22.7 & 23.4 & 21.3 \\ 
Null-text Inversion & 16.3 & 23.04 & 152.1 & 81.9 & 80.0 & 13.76 & 32.5 & 37.2 & 36.0 \\ 
Direct Inversion & 18.5 & 23.34 & 134.2 & 78.6 & 81.1 & 13.76 & 30.8 & 33.7 & 31.6 \\ 
Inversion-Free Editing & 14.6 & \textbf{24.11} & 96.8 & 80.1 & \textbf{82.9} & 14.23 & 45.6 & 47.9 & 43.7 \\ 
\hline 
Moodifier (Ours) & 17.4 & 22.19 & \textbf{94.9} & 127.3 & 82.2 & \textbf{16.13} & \textbf{66.7} & \textbf{73.3} & \textbf{80.0} \\ 
\hline 
\end{tabular} 
} 
\caption{Comparison of image editing methods using emotion-based descriptive prompts. Emotions were adapted to category-appropriate attributes for each domain (see details in the Supp.). $^*$CLIP score is measured via torchmetrics using pretrained OpenAI CLIP-ViT-Large-Patch14, not our MoodifyCLIP, to ensure unbiased evaluation. Human preference metrics represent hit rates (\%) from MTurk evaluations where participants selected their top 3 most-preferred models for each image, showing strong preference for Moodifier.
} 
\label{tab:comparison} 
\end{table*}

Quantitatively, results in Table \ref{tab:comparison} demonstrate this balance, with Moodifier achieving competitive performance on structural metrics while outperforming significantly in emotional accuracy measures.  Methods focusing on pixel-level fidelity (Plug-and-Play, Inversion-Free Editing) struggle with emotional expression, whereas those making bolder changes (Instruct-Diffusion) sacrifice structural integrity. Moodifier strikes an optimal balance, achieving the best LPIPS score (94.9) and CLIP similarity (16.13) while preserving structural integrity.
%maintaining reasonable structure.
Moreover, MTurk human evaluations confirm that users prioritize effective emotional transformation while valuing original integrity. The high overall preference score validates our balanced approach.

\begin{figure*}[!htbp]
    \centering
    \includegraphics[width=\textwidth]{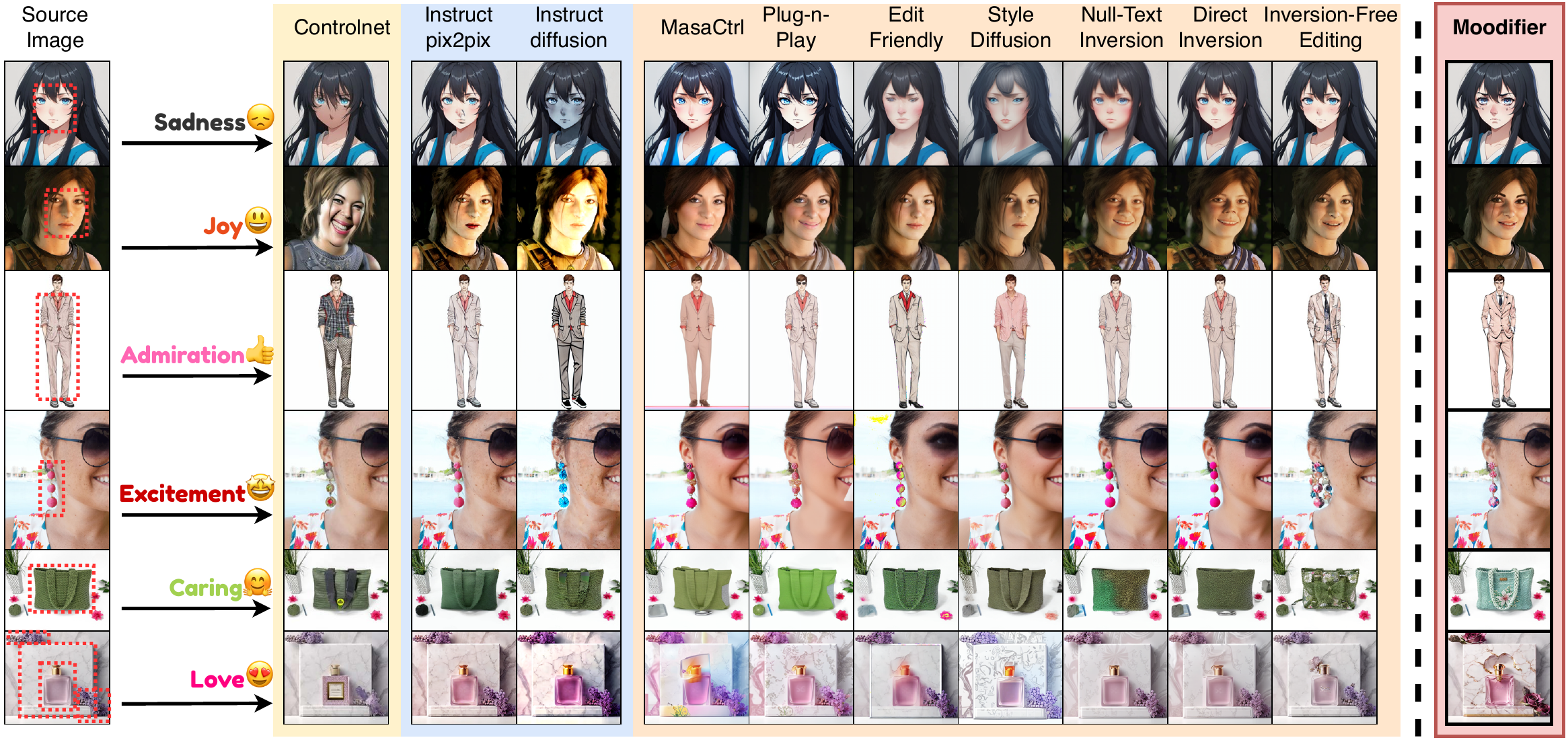}
    \caption{Qualitative comparisons with other editing methods. Zoom in for better view.}
    \label{fig:moodifier_comparisons}
    
\end{figure*}

Qualitatively, the comparisons in Fig. \ref{fig:moodifier_comparisons} show the limitations of competing methods, including identity change (ControlNet), artifacts (InstructPix2Pix, Instruct-Diffusion), weak emotional conveyance (StyleDiffusion, EditFriendly), or overly subtle modifications (MasaCtrl, Plug-n-Play). Moodifier, built upon MoodArchive and MoodifyCLIP, effectively addresses these limitations and it introduces meaningful emotional elements while maintaining strong structural integrity across diverse objects and emotions.

\begin{table*}[!htbp]
\centering
\resizebox{\textwidth}{!}{
\begin{tabular}{c|cc|c|cccc|c}
\cline{1-9}
\multirow{2}{*}{CLIP Model} & \multicolumn{2}{c|}{Edit Method} & \multicolumn{1}{c|}{Structure} & \multicolumn{4}{c|}{Background Preservation} & \multicolumn{1}{c}{CLIP Score} \\
\cline{2-9}
& Detail Prompts ($P_E$) & AttnMask ($M_E$)
& Distance$_{\times 10^3}$ $\downarrow$ & PSNR $\uparrow$ & LPIPS$_{\times 10^3}$ $\downarrow$ & MSE$_{\times 10^4}$ $\downarrow$ & SSIM$_{\times 10^2}$ $\uparrow$ & Similarity $\uparrow$ \\
\cline{1-9}
\multirow{4}{*}{CLIP-ViT-L-14} & $\times$ & $\times$ & 23.33 & 18.81 & 150.2 & 155.1 & 61.40 & 15.03 \\
& \checkmark & $\times$ & 22.30 & 19.01 & 146.2 & 152.3 & 69.32 & 15.80 \\
& $\times$ & \checkmark & 18.61 & 20.82 & 120.3 & 134.4 & 81.50 & 15.15\\
& \checkmark & \checkmark & \textbf{18.32} & 20.58 & 116.1 & 132.9 & \textbf{82.23} & 15.64 \\
\cline{1-9}
\multirow{2}{*}{MoodifyCLIP-ViT-L-14} & \checkmark & $\times$ & 21.90 & 19.91 & 148.7 & 147.6 & 65.61 & \textbf{16.07} \\
& \checkmark & \checkmark & $18.54^{\dagger}$ & \textbf{21.55} & \textbf{102.2} & \textbf{131.2} & $81.63^{\dagger}$ & $16.03^{\dagger}$ \\
\cline{1-9}
\end{tabular}
}
\caption{Ablation study of Moodifier variants. Note that no facial images are included in this study due to their fixed editing regions;
%, as their fixed editing regions provide limited room to show performance differences in ablation settings;
%Note: (1) no configurations without detailed prompts for Moodifier as it is designed to leverage rich emotional descriptions; (2) facial images not included, as their fixed editing regions offer limited opportunity to demonstrate Moodifier's capabilities with complex emotional transformations, 
this explains differences in the last row compared to Table \ref{tab:comparison}. Bold values indicate best performance; $^{\dagger}$ values indicate second-best performance.}
\label{tab:moodifier_ablation}
\end{table*}

\begin{figure}[!htbp]
    \centering
    \includegraphics[width=0.47\textwidth]{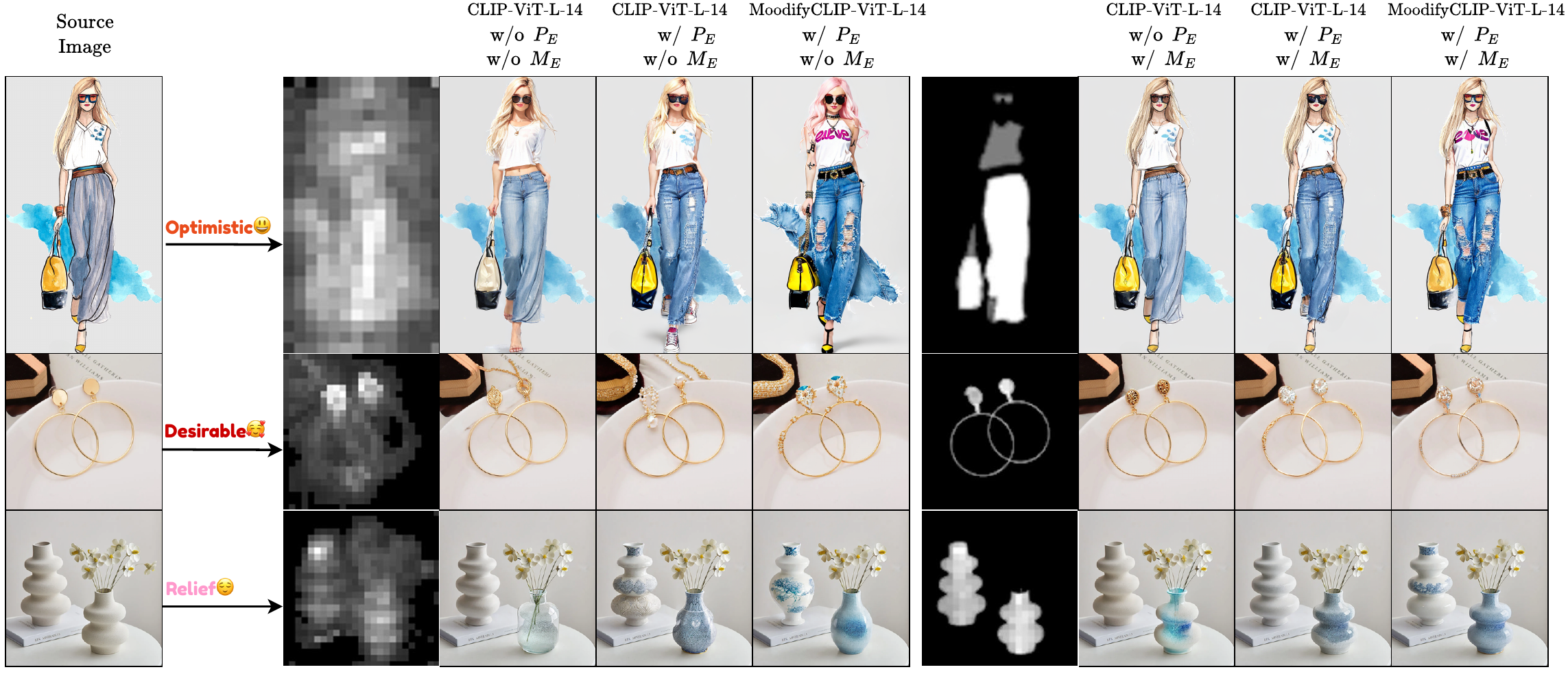}
    \caption{A qualitative example for ablation. Ours allows faithful and rich
semantic changes as well as better structural integrity perseveration. (See notation reference from Fig. \ref{fig:moodifer_workflow}, where $P_E$ represents the detailed MLLM-generated prompt and $M_E$ denotes the emotion stimulus masks). Zoom in for better view.}
    \label{fig:moodifier_ablations}    
\end{figure}

The ablation study (Fig. \ref{fig:moodifier_ablations}, Table \ref{tab:moodifier_ablation}) demonstrates that using prompts alone introduces emotional elements but sacrifices structural integrity, whereas using masks alone preserves structure but results in minimal emotional transformation. Combining both components yields optimal results,
confirming that emotional transformation requires guidance on both what to change (masks) and how to change it (emotion-rich prompts).

%% file: sec/5_conclusion.tex
\section{Conclusion and Discussion}

In conclusion, we introduce an integrated framework consisting of MoodArchive (a dataset of 8M+ images paired with emotion-relevant captions), MoodifyCLIP (a CLIP model specialized in the emotion dimension), and Moodifier (an MLLM-enhanced emotion-driven image editor). Our approach bridges abstract emotions and concrete visual changes while preserving content integrity. Extensive evaluations demonstrate our framework's ability to achieve high emotional accuracy while maintaining structural integrity, empowering creators to efficiently visualize emotional variations across diverse applications. 

In our future work, we will continue to increase the percentage of image-caption pairs in MoodArchive verified by human evaluators. We also recognize that the lack of standardized benchmarks for evaluating emotion-driven editing remains an open challenge. To address this, we plan to develop comprehensive datasets featuring images paired with their emotionally transformed versions, validated by human evaluators.

%% file: main.bbl
\begin{thebibliography}{10}

\bibitem{Zhang2023AddingCC}
Lvmin Zhang, Anyi Rao, and Maneesh Agrawala.
\newblock Adding conditional control to text-to-image diffusion models.
\newblock {\em 2023 IEEE/CVF International Conference on Computer Vision (ICCV)}, pages 3813--3824, 2023.

\bibitem{Brooks2022InstructPix2PixLT}
Tim Brooks, Aleksander Holynski, and Alexei~A. Efros.
\newblock Instructpix2pix: Learning to follow image editing instructions.
\newblock {\em 2023 IEEE/CVF Conference on Computer Vision and Pattern Recognition (CVPR)}, pages 18392--18402, 2022.

\bibitem{Geng23instructdiff}
Zigang Geng, Binxin Yang, Tiankai Hang, Chen Li, Shuyang Gu, Ting Zhang, Jianmin Bao, Zheng Zhang, Han Hu, Dong Chen, and Baining Guo.
\newblock Instructdiffusion: {A} generalist modeling interface for vision tasks.
\newblock In {\em Proc. CVPR}, 2024.

\bibitem{chen2015microsoft}
Xinlei Chen, Hao Fang, Tsung-Yi Lin, Ramakrishna Vedantam, Saurabh Gupta, Piotr Doll{\'a}r, and C~Lawrence Zitnick.
\newblock Microsoft coco captions: Data collection and evaluation server.
\newblock {\em arXiv preprint arXiv:1504.00325}, 2015.

\bibitem{young-etal-2014-image}
Peter Young, Alice Lai, Micah Hodosh, and Julia Hockenmaier.
\newblock From image descriptions to visual denotations: New similarity metrics for semantic inference over event descriptions.
\newblock {\em Transactions of the Association for Computational Linguistics}, 2:67--78, 2014.

\bibitem{ekman1987universals}
Paul Ekman, Wallace~V Friesen, Maureen O'sullivan, Anthony Chan, Irene Diacoyanni-Tarlatzis, Karl Heider, Rainer Krause, William~Ayhan LeCompte, Tom Pitcairn, Pio~E Ricci-Bitti, et~al.
\newblock Universals and cultural differences in the judgments of facial expressions of emotion.
\newblock {\em Journal of personality and social psychology}, 53(4):712, 1987.

\bibitem{machajdik2010affective}
Jana Machajdik and Allan Hanbury.
\newblock Affective image classification using features inspired by psychology and art theory.
\newblock In {\em Proceedings of the 18th ACM international conference on Multimedia}, pages 83--92, 2010.

\bibitem{peng2015mixed}
Kuan-Chuan Peng, Tsuhan Chen, Amir Sadovnik, and Andrew~C Gallagher.
\newblock A mixed bag of emotions: Model, predict, and transfer emotion distributions.
\newblock In {\em Proceedings of the IEEE conference on computer vision and pattern recognition}, pages 860--868, 2015.

\bibitem{yang2023emoset}
Jingyuan Yang, Qirui Huang, Tingting Ding, Dani Lischinski, Danny Cohen-Or, and Hui Huang.
\newblock Emoset: A large-scale visual emotion dataset with rich attributes.
\newblock In {\em Proceedings of the IEEE/CVF International Conference on Computer Vision}, pages 20383--20394, 2023.

\bibitem{kosti2019context}
Ronak Kosti, Jose~M Alvarez, Adria Recasens, and Agata Lapedriza.
\newblock Context based emotion recognition using emotic dataset.
\newblock {\em IEEE transactions on pattern analysis and machine intelligence}, 42(11):2755--2766, 2019.

\bibitem{achlioptas2023affection}
Panos Achlioptas, Maks Ovsjanikov, Leonidas Guibas, and Sergey Tulyakov.
\newblock Affection: Learning affective explanations for real-world visual data.
\newblock In {\em Proceedings of the IEEE/CVF Conference on Computer Vision and Pattern Recognition}, pages 6641--6651, 2023.

\bibitem{radford2021learning}
Alec Radford, Jong~Wook Kim, Chris Hallacy, Aditya Ramesh, Gabriel Goh, Sandhini Agarwal, Girish Sastry, Amanda Askell, Pamela Mishkin, Jack Clark, et~al.
\newblock Learning transferable visual models from natural language supervision.
\newblock In {\em International conference on machine learning}, pages 8748--8763. PMLR, 2021.

\bibitem{guopen}
Xiuye Gu, Tsung-Yi Lin, Weicheng Kuo, and Yin Cui.
\newblock Open-vocabulary object detection via vision and language knowledge distillation.
\newblock In {\em International Conference on Learning Representations}, 2023.

\bibitem{li2022grounded}
Liunian~Harold Li, Pengchuan Zhang, Haotian Zhang, Jianwei Yang, Chunyuan Li, Yiwu Zhong, Lijuan Wang, Lu~Yuan, Lei Zhang, Jenq-Neng Hwang, et~al.
\newblock Grounded language-image pre-training.
\newblock In {\em Proceedings of the IEEE/CVF conference on computer vision and pattern recognition}, pages 10965--10975, 2022.

\bibitem{lilanguage}
Boyi Li, Kilian~Q Weinberger, Serge Belongie, Vladlen Koltun, and Rene Ranftl.
\newblock Language-driven semantic segmentation.
\newblock In {\em International Conference on Learning Representations}, 2023.

\bibitem{xu2022groupvit}
Jiarui Xu, Shalini De~Mello, Sifei Liu, Wonmin Byeon, Thomas Breuel, Jan Kautz, and Xiaolong Wang.
\newblock Groupvit: Semantic segmentation emerges from text supervision.
\newblock In {\em Proceedings of the IEEE/CVF conference on computer vision and pattern recognition}, pages 18134--18144, 2022.

\bibitem{Luo2021CLIP4ClipAE}
Huaishao Luo, Lei Ji, Ming Zhong, Yang Chen, Wen Lei, Nan Duan, and Tianrui Li.
\newblock Clip4clip: An empirical study of clip for end to end video clip retrieval.
\newblock {\em Neurocomputing}, 508:293--304, 2021.

\bibitem{Xu2021VideoCLIPCP}
Hu~Xu, Gargi Ghosh, Po-Yao~(Bernie) Huang, Dmytro Okhonko, Armen Aghajanyan, and Florian Metze Luke Zettlemoyer~Christoph Feichtenhofer.
\newblock Videoclip: Contrastive pre-training for zero-shot video-text understanding.
\newblock In {\em Conference on Empirical Methods in Natural Language Processing}, 2021.

\bibitem{Crowson2022VQGANCLIPOD}
Katherine Crowson, Stella Biderman, Daniel Kornis, Dashiell Stander, Eric Hallahan, Louis Castricato, and Edward Raff.
\newblock Vqgan-clip: Open domain image generation and editing with natural language guidance.
\newblock In {\em European Conference on Computer Vision}, 2022.

\bibitem{frans2022clipdraw}
Kevin Frans, Lisa Soros, and Olaf Witkowski.
\newblock Clipdraw: Exploring text-to-drawing synthesis through language-image encoders.
\newblock {\em Advances in Neural Information Processing Systems}, 35:5207--5218, 2022.

\bibitem{Ramesh2022HierarchicalTI}
Aditya Ramesh, Prafulla Dhariwal, Alex Nichol, Casey Chu, and Mark Chen.
\newblock Hierarchical text-conditional image generation with clip latents.
\newblock {\em ArXiv}, abs/2204.06125, 2022.

\bibitem{vinker2022clipasso}
Yael Vinker, Ehsan Pajouheshgar, Jessica~Y Bo, Roman~Christian Bachmann, Amit~Haim Bermano, Daniel Cohen-Or, Amir Zamir, and Ariel Shamir.
\newblock Clipasso: Semantically-aware object sketching.
\newblock {\em ACM Transactions on Graphics (TOG)}, 41(4):1--11, 2022.

\bibitem{Kim2023RegionAwarePF}
Dahun Kim, Anelia Angelova, and Weicheng Kuo.
\newblock Region-aware pretraining for open-vocabulary object detection with vision transformers.
\newblock {\em 2023 IEEE/CVF Conference on Computer Vision and Pattern Recognition (CVPR)}, pages 11144--11154, 2023.

\bibitem{zeng2022multi}
Yan Zeng, Xinsong Zhang, and Hang Li.
\newblock Multi-grained vision language pre-training: Aligning texts with visual concepts.
\newblock In {\em International Conference on Machine Learning}, pages 25994--26009. PMLR, 2022.

\bibitem{fan2023improving}
Lijie Fan, Dilip Krishnan, Phillip Isola, Dina Katabi, and Yonglong Tian.
\newblock Improving clip training with language rewrites.
\newblock {\em Advances in Neural Information Processing Systems}, 36:35544--35575, 2023.

\bibitem{tian2023stablerep}
Yonglong Tian, Lijie Fan, Phillip Isola, Huiwen Chang, and Dilip Krishnan.
\newblock Stablerep: Synthetic images from text-to-image models make strong visual representation learners.
\newblock {\em Advances in Neural Information Processing Systems}, 36:48382--48402, 2023.

\bibitem{Yang2023ALIPAL}
Kaicheng Yang, Jiankang Deng, Xiang An, Jiawei Li, Ziyong Feng, Jia Guo, Jing Yang, and Tongliang Liu.
\newblock Alip: Adaptive language-image pre-training with synthetic caption.
\newblock {\em 2023 IEEE/CVF International Conference on Computer Vision (ICCV)}, pages 2910--2919, 2023.

\bibitem{pmlr-v202-zhao23l}
Liming Zhao, Kecheng Zheng, Yun Zheng, Deli Zhao, and Jingren Zhou.
\newblock {RLEG}: Vision-language representation learning with diffusion-based embedding generation.
\newblock In Andreas Krause, Emma Brunskill, Kyunghyun Cho, Barbara Engelhardt, Sivan Sabato, and Jonathan Scarlett, editors, {\em Proceedings of the 40th International Conference on Machine Learning}, volume 202 of {\em Proceedings of Machine Learning Research}, pages 42247--42258. PMLR, 23--29 Jul 2023.

\bibitem{Lai2023VeCLIPIC}
Zhengfeng Lai, Haotian Zhang, Wentao Wu, Haoping Bai, Aleksei Timofeev, Xianzhi Du, Zhe Gan, Jiulong Shan, Chen-Nee Chuah, Yinfei Yang, and Meng Cao.
\newblock Veclip: Improving clip training via visual-enriched captions.
\newblock In {\em European Conference on Computer Vision}, 2023.

\bibitem{liu2023mllms}
Yanqing Liu, Kai Wang, Wenqi Shao, Ping Luo, Yu~Qiao, Mike~Zheng Shou, Kaipeng Zhang, and Yang You.
\newblock Mllms-augmented visual-language representation learning.
\newblock {\em CoRR}, 2023.

\bibitem{Goodfellow2014GenerativeAN}
Ian~J. Goodfellow, Jean Pouget-Abadie, Mehdi Mirza, Bing Xu, David Warde-Farley, Sherjil Ozair, Aaron~C. Courville, and Yoshua Bengio.
\newblock Generative adversarial nets.
\newblock In {\em Neural Information Processing Systems}, 2014.

\bibitem{Reed2016GenerativeAT}
Scott~E. Reed, Zeynep Akata, Xinchen Yan, Lajanugen Logeswaran, Bernt Schiele, and Honglak Lee.
\newblock Generative adversarial text to image synthesis.
\newblock In {\em International Conference on Machine Learning}, 2016.

\bibitem{ho2020denoising}
Jonathan Ho, Ajay Jain, and Pieter Abbeel.
\newblock Denoising diffusion probabilistic models.
\newblock {\em Advances in neural information processing systems}, 33:6840--6851, 2020.

\bibitem{ramesh2022hierarchical}
Aditya Ramesh, Prafulla Dhariwal, Alex Nichol, Casey Chu, and Mark Chen.
\newblock Hierarchical text-conditional image generation with clip latents.
\newblock {\em arXiv preprint arXiv:2204.06125}, 2022.

\bibitem{saharia2022photorealistic}
Chitwan Saharia, William Chan, Saurabh Saxena, Lala Li, Jay Whang, Emily~L Denton, Kamyar Ghasemipour, Raphael Gontijo~Lopes, Burcu Karagol~Ayan, Tim Salimans, et~al.
\newblock Photorealistic text-to-image diffusion models with deep language understanding.
\newblock {\em Advances in neural information processing systems}, 35:36479--36494, 2022.

\bibitem{rombach2022high}
Robin Rombach, Andreas Blattmann, Dominik Lorenz, Patrick Esser, and Bj{\"o}rn Ommer.
\newblock High-resolution image synthesis with latent diffusion models.
\newblock In {\em Proceedings of the IEEE/CVF conference on computer vision and pattern recognition}, pages 10684--10695, 2022.

\bibitem{Meng2021SDEditGI}
Chenlin Meng, Yutong He, Yang Song, Jiaming Song, Jiajun Wu, Jun-Yan Zhu, and Stefano Ermon.
\newblock Sdedit: Guided image synthesis and editing with stochastic differential equations.
\newblock In {\em International Conference on Learning Representations}, 2021.

\bibitem{hertz2022prompt}
Amir Hertz, Ron Mokady, Jay Tenenbaum, Kfir Aberman, Yael Pritch, and Daniel Cohen-Or.
\newblock Prompt-to-prompt image editing with cross attention control.
\newblock {\em arXiv preprint arXiv:2208.01626}, 2022.

\bibitem{Kawar2022ImagicTR}
Bahjat Kawar, Shiran Zada, Oran Lang, Omer Tov, Hui-Tang Chang, Tali Dekel, Inbar Mosseri, and Michal Irani.
\newblock Imagic: Text-based real image editing with diffusion models.
\newblock {\em 2023 IEEE/CVF Conference on Computer Vision and Pattern Recognition (CVPR)}, pages 6007--6017, 2022.

\bibitem{gu2023photoswap}
Jing Gu, Yilin Wang, Nanxuan Zhao, Tsu-Jui Fu, Wei Xiong, Qing Liu, Zhifei Zhang, He~Zhang, Jianming Zhang, HyunJoon Jung, et~al.
\newblock Photoswap: Personalized subject swapping in images.
\newblock {\em Advances in Neural Information Processing Systems}, 36:35202--35217, 2023.

\bibitem{Nichol2021GLIDETP}
Alex Nichol, Prafulla Dhariwal, Aditya Ramesh, Pranav Shyam, Pamela Mishkin, Bob McGrew, Ilya Sutskever, and Mark Chen.
\newblock Glide: Towards photorealistic image generation and editing with text-guided diffusion models.
\newblock In {\em International Conference on Machine Learning}, 2021.

\bibitem{Avrahami2021BlendedDF}
Omri Avrahami, Dani Lischinski, and Ohad Fried.
\newblock Blended diffusion for text-driven editing of natural images.
\newblock {\em 2022 IEEE/CVF Conference on Computer Vision and Pattern Recognition (CVPR)}, pages 18187--18197, 2021.

\bibitem{Wang2022ImagenEA}
Su~Wang, Chitwan Saharia, Ceslee Montgomery, Jordi Pont-Tuset, Shai Noy, Stefano Pellegrini, Yasumasa Onoe, Sarah Laszlo, David~J. Fleet, Radu Soricut, Jason Baldridge, Mohammad Norouzi, Peter Anderson, and William Chan.
\newblock Imagen editor and editbench: Advancing and evaluating text-guided image inpainting.
\newblock {\em 2023 IEEE/CVF Conference on Computer Vision and Pattern Recognition (CVPR)}, pages 18359--18369, 2022.

\bibitem{bar2022text2live}
Omer Bar-Tal, Dolev Ofri-Amar, Rafail Fridman, Yoni Kasten, and Tali Dekel.
\newblock Text2live: Text-driven layered image and video editing.
\newblock In {\em European conference on computer vision}, pages 707--723. Springer, 2022.

\bibitem{couairon2023diffedit}
Guillaume Couairon, Jakob Verbeek, Holger Schwenk, and Matthieu Cord.
\newblock Diffedit: Diffusion-based semantic image editing with mask guidance.
\newblock In {\em ICLR 2023 (Eleventh International Conference on Learning Representations)}, 2023.

\bibitem{Tumanyan2022PlugandPlayDF}
Narek Tumanyan, Michal Geyer, Shai Bagon, and Tali Dekel.
\newblock Plug-and-play diffusion features for text-driven image-to-image translation.
\newblock {\em 2023 IEEE/CVF Conference on Computer Vision and Pattern Recognition (CVPR)}, pages 1921--1930, 2022.

\bibitem{Cao2023MasaCtrlTM}
Ming Cao, Xintao Wang, Zhongang Qi, Ying Shan, Xiaohu Qie, and Yinqiang Zheng.
\newblock Masactrl: Tuning-free mutual self-attention control for consistent image synthesis and editing.
\newblock {\em 2023 IEEE/CVF International Conference on Computer Vision (ICCV)}, pages 22503--22513, 2023.

\bibitem{demszky2020goemotions}
Dorottya Demszky, Dana Movshovitz-Attias, Jeongwoo Ko, Alan Cowen, Gaurav Nemade, and Sujith Ravi.
\newblock Goemotions: A dataset of fine-grained emotions.
\newblock {\em arXiv preprint arXiv:2005.00547}, 2020.

\bibitem{liu2024visual}
Haotian Liu, Chunyuan Li, Qingyang Wu, and Yong~Jae Lee.
\newblock Visual instruction tuning.
\newblock {\em Advances in neural information processing systems}, 36, 2024.

\bibitem{Zhang2024LongCLIPUT}
Beichen Zhang, Pan Zhang, Xiao wen Dong, Yuhang Zang, and Jiaqi Wang.
\newblock Long-clip: Unlocking the long-text capability of clip.
\newblock In {\em European Conference on Computer Vision}, 2024.

\bibitem{shi2024ot}
Liangliang Shi, Jack Fan, and Junchi Yan.
\newblock Ot-clip: Understanding and generalizing clip via optimal transport.
\newblock In {\em Forty-first International Conference on Machine Learning}, 2024.

\bibitem{xu2023inversion}
Sihan Xu, Yidong Huang, Jiayi Pan, Ziqiao Ma, and Joyce Chai.
\newblock Inversion-free image editing with natural language.
\newblock In {\em Conference on Computer Vision and Pattern Recognition 2024}, 2024.

\bibitem{mathews2016senticap}
Alexander Mathews, Lexing Xie, and Xuming He.
\newblock Senticap: Generating image descriptions with sentiments.
\newblock In {\em Proceedings of the AAAI conference on artificial intelligence}, volume~30, 2016.

\bibitem{Achlioptas2022AffectionLA}
Panos Achlioptas, Maks Ovsjanikov, Leonidas~J. Guibas, and S.~Tulyakov.
\newblock Affection: Learning affective explanations for real-world visual data.
\newblock {\em 2023 IEEE/CVF Conference on Computer Vision and Pattern Recognition (CVPR)}, pages 6641--6651, 2022.

\bibitem{HubermanSpiegelglas2023AnEF}
Inbar Huberman-Spiegelglas, Vladimir~B. Kulikov, and Tomer Michaeli.
\newblock An edit friendly ddpm noise space: Inversion and manipulations.
\newblock {\em 2024 IEEE/CVF Conference on Computer Vision and Pattern Recognition (CVPR)}, pages 12469--12478, 2023.

\bibitem{Wang2023StyleDiffusionCD}
Zhizhong Wang, Lei Zhao, and Wei Xing.
\newblock Stylediffusion: Controllable disentangled style transfer via diffusion models.
\newblock {\em 2023 IEEE/CVF International Conference on Computer Vision (ICCV)}, pages 7643--7655, 2023.

\bibitem{mokady2023null}
Ron Mokady, Amir Hertz, Kfir Aberman, Yael Pritch, and Daniel Cohen-Or.
\newblock Null-text inversion for editing real images using guided diffusion models.
\newblock In {\em Proceedings of the IEEE/CVF Conference on Computer Vision and Pattern Recognition}, pages 6038--6047, 2023.

\bibitem{ju2024pnp}
Xuan Ju, Ailing Zeng, Yuxuan Bian, Shaoteng Liu, and Qiang Xu.
\newblock Pnp inversion: Boosting diffusion-based editing with 3 lines of code.
\newblock In {\em The Twelfth International Conference on Learning Representations}, 2024.

\end{thebibliography}
